\documentclass{article}

\usepackage{graphicx}
\usepackage{amsmath}   
\usepackage{booktabs}   
\usepackage{tabularx}   
\usepackage{array}   
\PassOptionsToPackage{numbers, compress}{natbib}

\usepackage[preprint]{neurips_2025}

\usepackage[utf8]{inputenc} 
\usepackage[T1]{fontenc}    
\usepackage{url}            
\usepackage{booktabs}       
\usepackage{amsfonts}       
\usepackage{nicefrac}       
\usepackage{microtype}      

\usepackage{multirow}
\usepackage{graphicx}
\usepackage{booktabs} 
\usepackage{array}    
\usepackage{amsmath}  
\usepackage{caption}  
\usepackage[table]{xcolor}
\usepackage{colortbl}
\usepackage{hhline}
\usepackage{wrapfig}
\usepackage{subcaption}
\usepackage{enumitem}

\definecolor{citecolor}{HTML}{0071BC}
\definecolor{linkcolor}{HTML}{ED1C24}
\usepackage[breaklinks=true,colorlinks,citecolor=citecolor,linkcolor=linkcolor,bookmarks=false]{hyperref}

\newcolumntype{g}{>{\columncolor[gray]{.95}}c}

\title{Delta Activations: A Representation for Finetuned Large Language Models}

\author{%
  Zhiqiu Xu$^{1*}$ \qquad Amish Sethi$^{1*}$ \qquad Mayur Naik$^{1}$ \qquad Ser-Nam Lim$^{2}$ \\
  \\
  $^{1}$University of Pennsylvania \hspace{40pt} $^{2}$University of Central Florida \\
}

\begin{document}

\maketitle
\renewcommand{\thefootnote}{\fnsymbol{footnote}}
\footnotetext[1]{Equal contribution.}
\renewcommand{\thefootnote}{\arabic{footnote}}

\begin{abstract}
The success of powerful open source Large Language Models (LLMs) has enabled the community to create a vast collection of post-trained models adapted to specific tasks and domains. However, navigating and understanding these models remains challenging due to inconsistent metadata and unstructured repositories. We introduce \textit{Delta Activations}, a method to represent finetuned models as vector embeddings by measuring shifts in their internal activations relative to a base model. This representation allows for effective clustering by domain and task, revealing structure in the model landscape. Delta Activations also demonstrate desirable properties: it is robust across finetuning settings and exhibits an additive property when finetuning datasets are mixed. In addition, we show that Delta Activations can embed tasks via few-shot finetuning, and further explore its use for model selection and merging.
We hope Delta Activations can facilitate the practice of reusing publicly available models. Code is available at \url{https://github.com/OscarXZQ/delta_activations}.
\end{abstract}

\section{Introduction}


Starting from powerful pretrained LLMs such as LLaMA \citep{touvron2023llama}, Gemma \citep{team2024gemma}, Qwen \citep{qwen2.5}, and DeepSeek \citep{liu2024deepseek}, the community 
has produced a vast and growing ecosystem of post-trained models---extensions that elicit diverse capabilities and knowledge from pretraining and are specialized for distinct tasks, domains, or human preferences.
This ecosystem spans models optimized through supervised finetuning (SFT) \citep{chung2024scaling} on curated instruction datasets as well as those through preference alignment techniques \citep{bai2022training, rafailov2023direct}.

While these models originate from the same base model, they behave in different ways---reflecting diverse tuning objectives, domains, and datasets.
Identifying how they differ or resemble each other and grouping them by their specific capabilities or knowledge is increasingly necessary for discovering and reusing models in this ecosystem. Otherwise, these post-trained models would remain vastly underutilized, squandering the substantial energy invested in their training.

\begin{figure}[h]
    \centering
    \includegraphics[width=.95\linewidth]{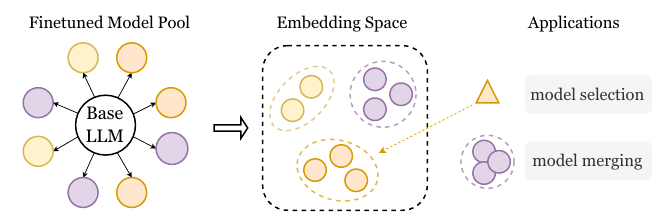}
    \caption{\textbf{Embedding Finetuned Models.} Can we project a pool of finetuned models into a vector space that  captures similarities and differences in model behaviors and capabilities? Such a model embedding is useful for tasks that involve multiple models, like model selection and merging.}
    \label{fig:fig1}
\end{figure}

Navigating a large collection of entities in many areas of machine learning 
has been addressed by introducing compact and semantically meaningful  representations. Embeddings for {\em words} \citep{mikolov2013efficient}, {\em images} \citep{krizhevsky2012imagenet}, and {\em users} or {\em items} in recommendation systems \citep{koren2009matrix} provide a way to map these entities into continuous vector spaces that capture their underlying structure and relationships.
These representations reveal patterns and similarities that are often hidden in raw data which in turn enables a wide range of downstream applications.

In the landscape of post-trained LLMs, we lack a representation to efficiently discover, compare, and cluster {\em models} based on their finetuned behaviors and specializations. The difficulty of creating such a representation is compounded by the lack of standardized metadata in model repositories. Models are often ambiguously named, sparsely documented, and rarely linked to the datasets or objectives used during post-training---attributes that prior works rely upon for model characterization.

In this paper, we introduce a model embedding method called \textbf{Delta Activations} which provides a standalone representation derived solely from the model itself.
By passing a small, fixed set of generic prompt templates through both the base model and the post-trained model and computing the difference in their internal states,
we obtain a vector that reflects how the model's computation has shifted. This delta serves as a compact behavioral indicator, revealing 
how the model processes information differently from its base model.

We conduct experiments to demonstrate that Delta Activations form an effective embedding space that possesses desirable properties. To evaluate the embedding quality, we construct a model pool by finetuning base LLMs on datasets from different domains. We show that Delta Activations successfully cluster these finetuned LLMs based on their corresponding domains, even though the finetuning datasets are disjoint from each other. Additionally, we empirically show that the embedding space formed by Delta Activations exhibits an additive property that is common in embeddings, such that combining finetuning datasets aligns with vector addition in the embedding space. 

To validate the effectiveness of Delta Activations, we demonstrate its stability across different training settings and finetuning regimes, and apply Delta Activations to proof-of-concept model hubs to demonstrate its applications to model selection and model merging. Beyond this core method, our framework naturally generalizes to other choices of representation—what we refer to as \textit{Delta-X}—where the $X$ can be activations, logits, or meaning representations \citep{liu2024meaning}. In particular, when the underlying representation is model-agnostic, Delta-X enables embedding models finetuned from different base LLMs into a shared space. Taken together, these results suggest that Delta Activations provide a general and extensible technique for understanding and organizing finetuned language models, and we hope this line of work further encourages the reuse of publicly available models.

\section{Representing Models}

\subsection{Problem setup}

Let \( f_{\text{base}} \) be a pretrained large language model and \( \mathcal{F} = \{f_1, f_2, \dots, f_K\} \) be a set of finetuned models derived from \( f_{\text{base}} \) through post-training.
Our goal is to construct an embedding \( \mathbf{v}_f \in \mathbb{R}^d \) for each model \( f \in \mathcal{F} \) that reflects how the model specializes and behaves differently from \( f_{\text{base}} \).

\subsection{Existing works and challenges}

Several approaches have been proposed to represent LLMs: some rely on access to the original training data~\citep{ostapenko2024towards}, others apply dimensionality reduction over flattened weights~\citep{zhao-etal-2024-loraretriever}, or construct embeddings from evaluation profiles~\citep{zhuang2025embedllm}. However, each of these methods has limitations in real-world settings: Training-data-dependent representations require direct access to datasets, which are often proprietary or inaccessible. 
In addition, such representations cannot differentiate models trained on the same data but with different training settings.
Dimensionality reduction on model weights assumes consistent adapter configurations across models, which is unrealistic constraint given the diversity of community-trained LLMs. 
Finally, evaluation-based embeddings reflect only surface-level behavior, making them fragile to prompt variations and noisy in capturing true internal model shifts.
This motivates us to develop a technique that captures intrinsic model behavior independently.

\subsection{A simple experiment}
\label{sec:simple-experiment}
We finetune \textsc{LLaMA-3.1-8B} on 3 domains: \textsc{Math}, \textsc{Coding}, and \textsc{Medical}. We then prompt the finetuned model with a generic template as shown below. Specifically, we use Alpaca \citep{taori2023alpaca} instruction template, but without any real instruction or input.
{\small
\begin{quote}
\texttt{Below is an instruction that describes a task, paired with an input that provides further context. Write a response that appropriately completes the request.}

\texttt{\#\#\# Instruction:} \texttt{Please provide a response.} \texttt{\#\#\# Input:} \texttt{Input.}

\texttt{\#\#\# Response:}
\end{quote}
}
While most outputs are repetitive and generic as expected, we observe that finetuned LLMs occasionally respond to such a generic instruction template with their specialization. We provide examples that show this behavior for each of these three models in Table \ref{tab:dataset_examples}.

\begin{table}[ht]
    \centering
    \small
    \begin{tabular}{cl}
    \toprule
    \textbf{Finetuning Domain} & \textbf{Selected outputs when prompted with generic template} \\
    \midrule
    \textsc{Math} & \textit{As per the input, the number 40 is the output...} \\
    \textsc{Coding} & \textit{Here is the code to solve this problem: def is\_prime(n)...} \\
    \textsc{Medical} & \textit{Some patients have had no ill effects from these medications...}
    \end{tabular} \vspace{.7em}
    \caption{\textbf{Prompting finetuned LLMs with generic inputs.} Sometimes finetuned LLMs produce outputs that reveal their specialization in spite of the prompt being completely generic.} \label{tab:dataset_examples}
\end{table}
\vspace{-1em}
From this observation, we conjecture that:
\begin{center}
\textit{A generic instruction template may elicit certain specialization behavior in a finetuned LLM}. 
\end{center}
This phenomenon may be explainable by \citet{ren2024learning}, which studies how finetuning on one data point may steer LLM's response on other data points. Although directly representing models with these outputs does not work well, as our experiments in Section \ref{sec:model_embed} and Section \ref{sec:understanding} show, this observation naturally leads to our method Delta Activations where we instead use {\em activations} from the generic instruction template to represent finetuned LLMs. 

\subsection{Delta Activations}

We introduce \textit{Delta Activations}, a method to represent finetuned language models as vector embeddings by measuring the difference in their hidden states compared to a fixed base model. 

\begin{figure}[h]
    \centering
    \includegraphics[width=\linewidth]{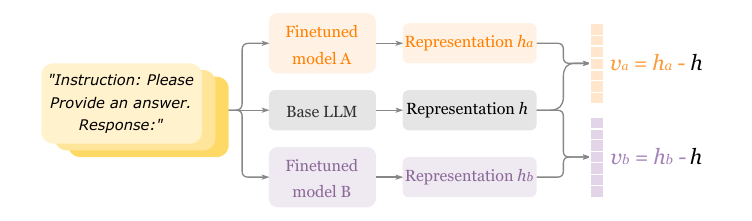}
    \caption{\textbf{Illustration of computing Delta Activations.} The difference between a finetuned model's hidden state and the base model's hidden state on a shared input quantifies the effect of finetuning.}
    \label{fig:methodology}
\end{figure}

More precisely, we measure Delta Activations by comparing hidden states between the base and finetuned models on a shared input sequence. Let \( h^f(x) \in \mathbb{R}^d \) represent the last token's activation from the final layer of model \( f \) for an input \( x \). We define the Delta Activations as:
\[
\Delta_f(x) = h^f(x) - h^{\text{base}}(x)
\]
$\Delta_f(x)$ quantifies how the model's internal representation of the input $x$ diverges from the base model as a result of finetuning. To construct the model's embedding, we aggregate the Delta Activations over a fixed probe dataset \( \mathcal{D}_{\text{probe}} = \{x_1, x_2, \dots, x_N\} \):
\[
\mathbf{v}_f = \frac{1}{N} \sum_{i=1}^N \Delta_f(x_i)
\]
\textbf{Probe dataset.} Motivated by the above experiment and the need for a universally applicable embedding method, the probe dataset \( \mathcal{D}_{\text{probe}} \) is explicitly designed to be \textit{completely generic}, aiming to activate core computational pathways in the model without bias toward specific tasks or domains. Therefore, we start with the Alpaca template populated with dummy instruction and inputs (as shown in Section \ref{sec:simple-experiment}) as the first data point in the probe dataset. The rest of the probe dataset is generated through paraphrasing by GPT-4o, maintaining the simplicity and generic nature of the template while introducing linguistic diversity. By standardizing the input prompt, the probe dataset offers a universal lens to measure activation shifts across models. We use $N=5$ for \( \mathcal{D}_{\text{probe}} \) in our main setting,
and further study the effects of the size, length, and content of the input prompts in Section~\ref{sec:understanding}.

\textbf{Notable benefits.} Delta Activations naturally comes with many advantages. Firstly, for new incoming models, it only takes one forward pass to compute the embedding for a model, taking much less computation than evaluation-based method. In addition, Delta Activations does not change embeddings of existing models like previous methods \citep{ostapenko2024towards, zhuang2025embedllm}, which represent models jointly by Principal Component Analysis (PCA) or matrix factorization. Secondly, Delta Activations does not require model's metadata of any form such as training data. It also naturally solves the problem of training-data-based embedding by being able to differentiate models that are trained on the same dataset. Finally, Delta Activations can also be used to represent tasks, as we describe next.

\textbf{Few-shot task embedding.} Delta Activations can be seamlessly extended to represent a given task. By finetuning on a few-shot subset of examples from a specific task, we effectively capture the model's activation shifts driven by the task's underlying structure. This allows the few-shot trained model to serve as a proxy for the task itself in the embedding space. Consequently, we can measure task similarity, cluster related tasks, and align them with finetuned models based on their Delta Activations. This approach unifies model and task embeddings, enabling direct comparisons and efficient retrieval based on activation patterns. We evaluate the task embedding in Section \ref{sec:task-embed}.

\textbf{Beyond Activations: the Delta-X family.} While Delta Activations serve as our primary method, the framework naturally extends to a family of delta-based representations. Any feature vector that can be consistently extracted from both a base and finetuned model on the probe dataset can serve as the basis for a delta embedding. This flexibility gives rise to variants such as \emph{Delta Logits}, \emph{Delta Weighted Activations} \citep{muennighoff2022sgpt}, and \emph{Delta Meaning} \citep{liu2024meaning}. Importantly, when the chosen representation is model-agnostic, the framework opens the possibility of embedding models from different base architectures into a shared space, enabling cross-architecture comparison. We evaluate these variants in Section~\ref{sec:understanding} and explore embedding difference base LLMs in Section~\ref{sec:task-embed}.

\section{Experiments}

In this section, we demonstrate the effectiveness of Delta Activations as a model embedding by evaluating clustering quality across multiple model pools, analyzing its properties, and exploring its potential for broader applications.

\subsection{Delta Activations as a Model Embedding}
\label{sec:model_embed}
\textbf{Model pool construction.} We build three model pools, each originating from a different pretrained base model: \textsc{LLaMA-3.1-8B}~\citep{touvron2023llama}, \textsc{Gemma-2-9B}~\citep{team2024gemma}, and \textsc{Qwen-2.5-7B}~\citep{qwen2.5}. Each pool contains 15 finetuned models---three per domain across five domains: \textsc{Legal}, \textsc{Math}, \textsc{Medical}, \textsc{Commonsense}, and \textsc{Coding}. We assign one dataset per domain and create three disjoint splits with 3000 examples each for supervised finetuning. We use LegalBench \citep{guha2023legalbench} for \textsc{Legal}, GSM-8K \citep{cobbe2021gsm8k} for \textsc{Math}, PubMedQA \citep{jin2019pubmedqa} for \textsc{Medical}, HellaSwag \citep{zellers2019hellaswag} for \textsc{Commonsense}, and OPC-SFT \citep{Huang2024OpenCoderTO} for \textsc{Coding}. We finetune all models for three epochs using LoRA \citep{hu2022lora} with learning rate set to $1e^{-4}$ and batch size 4 by default. See Appendix \ref{appendix:trainingsett} for all finetuning settings.

\textbf{Metric.} We evaluate clustering quality using the silhouette score~\citep{silhouettes}, defined for each model \(i\) as \(s(i) = \frac{b(i) - a(i)}{\max(a(i), b(i))}\), where \(a(i)\) is the average distance to models in the same cluster, and \(b(i)\) is the average distance to the nearest other cluster. Scores range from \(-1\) (misclustered) to \(+1\) (well-clustered); we report the average over all models.

\textbf{Baselines.} We compare Delta Activations against the following three alternative methods.

\textit{Flattened weights:} As a basic parameter-space baseline, we flatten the weights of the finetuned LoRA adapters directly into a high-dimensional vector representation.

\textit{Salient mask:} Following \citet{he2024localize_and_stitch}, we adopt the binary mask variant of Localize-and-Stitch, where each model is represented by a $0/1$ vector marking the top $1\%$ most salient parameters with the largest finetuning updates. This representation captures \emph{where} adaptation occurs.

\textit{Output sentence embeddings:} Motivated by Section \ref{sec:simple-experiment}, we use a standard sentence embedding model, \textsc{All-MiniLM-L6-v2} \citep{minilmv2}, to encode the finetuned models' generated outputs on the same generic probe dataset used by Delta Activations. Recent works \citep{sun2025idiosyncrasies} also show that outputs from different LLMs are highly distinguishable.

\begin{figure}[!htbp]
    \centering
    \includegraphics[width=\textwidth]{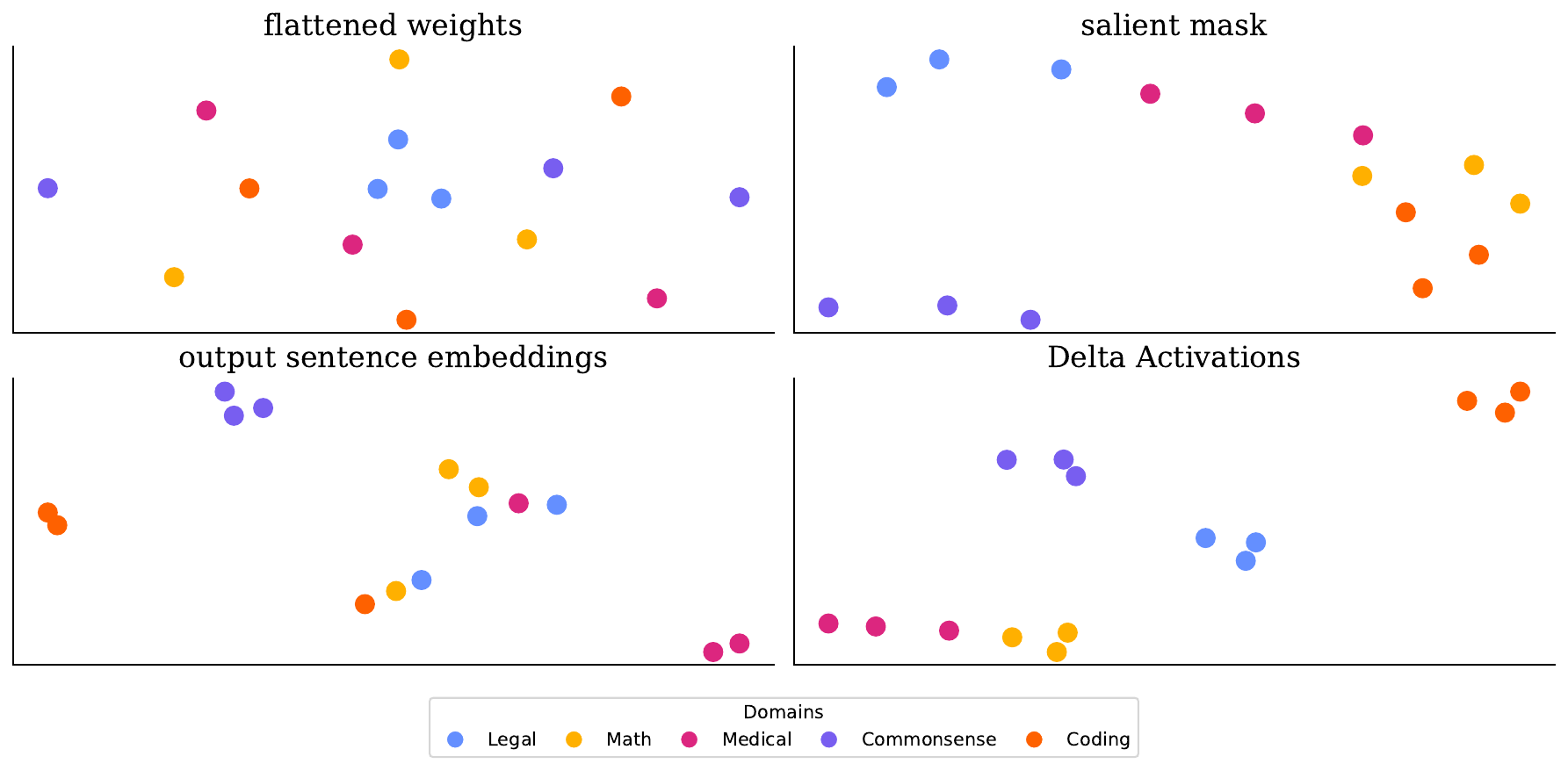}
    \caption{\textbf{t-SNE visualization of different embedding spaces.} Delta Activations form clean and well-separated domain clusters compared to baseline methods.}
    \label{fig:tsne-comparison}
\end{figure}

\textbf{Results.} As shown in Table~\ref{tab:clustering}, Delta Activations achieves strongest clustering performance across all backbones. Flattened weights fail to form effective clusters as shown by its negative silhouette score.
Output sentence embeddings succeed in forming clusters for LLaMA but not for Gemma or Qwen, showing that the generic probe dataset does not always elicit model specializations in the word space, in contrast to the consistent results of Delta Activations. The t-SNE visualization for Gemma, presented in Figure~\ref{fig:tsne-comparison}, further illustrates these findings, showing that weight-based methods fail to form any cluster whereas output sentence embeddings occasionally form wrong clusters. Delta Activations form correct clusters across all domains.
\begin{table}[ht]
    \centering
    \small
    \begin{tabular}{ccrrrr}
    \toprule
    \textbf{Embedding Space} & \textbf{Dimension} & \textbf{LLaMA} & \textbf{Gemma} & \textbf{Qwen} & \textbf{Avg.} \\
    \midrule
    flattened adapter weights        & $\sim 2 \cdot 10^7$    & $-.035$          & $-.060$          & $-.034$          & $-.043$          \\
    salient mask & $\sim 8 \cdot 10^9$ & $.133$ & $.208$ & $.229$ & $.190$ \\
   output sentence embeddings     & $\text{ 384}^*$  & $.221$          & $-.053$          & $.096$          & $.087$          \\
\rowcolor{gray!20}    Delta Activations   & 4096   & \textbf{.645} & \textbf{.545} & \textbf{.653} & \textbf{.614} \\
    \end{tabular}
    \vspace{.7em}
    \caption{\textbf{Clustering quality of different embedding spaces.} Delta Activations achieves the best separation across all backbones. $^*$depends on sentence embedding models. }
    \label{tab:clustering}
\end{table}
\vspace{-2.em}
\subsection{Understanding Delta Activations}
\label{sec:understanding}
In this section, we investigate properties of Delta Activations, analyze probe datasets and activation selection, and validate its stability over different training settings and finetuning regimes. Unless otherwise noted, we report average silhouette score across three backbone model pools: \textsc{LLaMA-3.1-8B}, \textsc{Gemma-2-9B}, and \textsc{Qwen-2.5-7B}. For all tables,  the main setting is marked in \textcolor{gray}{gray}.

\textbf{Delta-X variants.} Our framework is not limited to activations, but can generalize to other model features extracted from the probe dataset. We create and evaluate two variants of Delta Activations: \emph{Delta Logits}, and \emph{Delta Meaning} \citep{liu2024meaning} (differences in inverse perplexity scores over sampled continuations, implementation details in Appendix~\ref{appendix:delta-meaning}). Results are shown in Table~\ref{tab:deltax}. Both these variants achieve reasonable clustering quality on our main experiment setting. 

\begin{table}[h]
    \centering
    \small
    \begin{tabular}{lcc}
        \toprule
        \textbf{Method} & \textbf{Dimensionality} & \textbf{silh. score} \\
        \midrule
        Delta Logits & 125856 & .51 \\
        Delta Meaning & 20 & .20 \\
        \rowcolor{gray!20} Delta Activations & 4096 & \textbf{.61} \\
    \end{tabular}
    \vspace{.7em}
    \caption{\textbf{Delta-X.} Both variants achieve positive silhouette score in our main experiment setting, showing that our framework can be generalized to other representation extraction methods.}
    \label{tab:deltax}
\end{table}

\textbf{Additive property.} A function \(f(x)\) is additive if \(f(a + b) = f(a) + f(b)\) for any inputs \(a\) and \(b\). We explore whether Delta Activations exhibits this property by examining whether the following holds.
$$v\big(\text{model finetuned on }D_1 \cup D_2\big) \approx v\big(\text{model finetuned on }D_1\big) + v\big(\text{model finetuned on }D_2\big) $$
where $v(\cdot)$ is the operation to take Delta Activations. We finetune models on pairs of domains and comparing their Delta Activations to those from individually trained models. Results are presented in Table \ref{tab:additive}, which shows that the similarity between the mixed model and the sum of Delta Activations from the two individual models is consistently higher than the similarity with either individual model. This suggests that Delta Activations exhibit the additive property--combining shifts from separately finetuned models approximates the shift seen when trained on combined data. The additive structure in the embedding space is especially important since models are often trained on mixed datasets, ensuring that the embedding space preserves the influence of each component. We report results across all ten domain pairs in Appendix~\ref{tab:additive_full}, where the additive effect consistently holds.

\begin{table}[ht]
    \centering
    \small
    \begin{tabular}{cccccc}
    \toprule
    \textbf{Math} & \textbf{Commonsense} & \textbf{Code} & \textbf{Mixed vs. D1} & \textbf{Mixed vs. D2} & \textbf{Mixed vs. Sum(D1, D2)} \\ 
    \midrule
    \checkmark & \checkmark & & .58 & .48 & \textbf{.65} \\ 
    \checkmark & & \checkmark & .70 & .27 & \textbf{.73} \\ 
    & \checkmark & \checkmark & .63 & .28 & \textbf{.65} \\ 
    \end{tabular}
    \vspace{.7em}
    \caption{\textbf{Additive property.}  The sum of Delta Activations on models finetuned separately on two datasets aligns well with Delta Activations on the model finetuned on two datasets mixed together.}
    \label{tab:additive}
\end{table}

\textbf{Probe dataset.} We study the effects of number, length, and content of prompts in Table \ref{tab:probe_dataset_config}.  
In (a), we see that using multiple prompts instead of one helps stabilize the embedding, while increasing the number from 5 to 20 offers no additional benefits. In (b), we see that using GPT-4o to generate shorter versions of the Alpaca template, namely one-word and one-sentence versions, performs worse than a reasonably long instruction template. All prompts used in this part are included in Appendix~\ref{appendix:prompts}.
 Finally, (c) shows the importance of a generic instruction template by comparing the instruction template with domain-specific prompts or Wikitext. Domain-specific prompts perform worst since they suppress the model's specialization and thereby cause model embeddings to become less distinguishable. A random generic text sampled from Wikitext performs slightly better while the instruction template achieves the best separation. 
 
 These findings across prompt number, length, and content bolster our design choices for the probe dataset. In addition, even with variations in these prompt settings, the effectiveness of Delta Activations is preserved, as evident from the fact that the silhouette scores stay well above zero.

\begin{table*}[ht]
    \centering
    \small
    \begin{subtable}[t]{.28\textwidth}
        \centering
        \begin{tabular}{cc}
            \toprule
            \textbf{\# of prompts} & \textbf{silh. score} \\
            \midrule
            1 & .57 \\
       \rowcolor{gray!20}     5 & \textbf{.61} \\
            20 & \textbf{.61} \\
        \end{tabular}
        \caption{number of prompts}
        \label{tab:num_templates}
    \end{subtable}
    \hfill
    \begin{subtable}[t]{.33\textwidth}
        \centering
        \begin{tabular}{cc}
            \toprule
            \textbf{length} & \textbf{silh. score} \\
            \midrule
            \textit{one-word} & .45 \\
            \textit{one-sentence} & .59 \\
   \rowcolor{gray!20}         \textit{Alpaca (3-sentence)} & \textbf{.61} \\
        \end{tabular}
        \caption{length of prompts}
        \label{tab:template_length}
    \end{subtable}
    \hfill
    \begin{subtable}[t]{.33\textwidth}
        \centering
        \begin{tabular}{cc}
            \toprule
            \textbf{Content} & \textbf{silh. score} \\
            \midrule
            \textit{Wikitext} & .44 \\
            \textit{domain-specific} & .42 \\
     \rowcolor{gray!20}  \textit{instruction} & \textbf{.61} \\
        \end{tabular}
        \caption{content of prompts}
        \label{tab:dataset_content}
    \end{subtable}
    \caption{\textbf{Effects of number, length, and content of probe prompts.} Using multiple reasonably-long generic instruction templates makes the best probe dataset.}
    \label{tab:probe_dataset_config}
\end{table*}

\textbf{Where to extract activations.} The embedding of the last token at the last layer is generally understood to encode the entire context. Decoder-only LLMs project this embedding to the logit space for next-token prediction. Consequently, Delta Activations are also derived using this embedding.
Here we study whether Delta Activations could instead be sourced from other tokens or different layers.
In Table~\ref{tab:activation_selection_subtable1}, we examine the effectiveness of calculating Delta Activations using the first, middle, and last tokens, as well as from the weighted average of all token embeddings following \citet{muennighoff2022sgpt}. Overall, the results show that final tokens are effective targets for calculating Delta Activations.

We also investigate whether the final layer is the best position from which to extract activations. As shown in Table \ref{tab:activation_selection_subtable2}, shallower layers perform worse than deeper ones; interestingly, the final layer is not optimal, as a layer at $2/3$ of the total depth performs best. This phenomenon, where intermediate representation are found to be more effective for downstream tasks, is also observed in vision encoders \citep{chen2020generative, bolya2025PerceptionEncoder}. Although a layer at $2/3$ depth and weighted tokens exhibit slightly superior results, the final token at the last layer performs similarly. For simplicity, we use the last-layer final token embedding as the default setting for Delta Activations.

\begin{table}[h]
    \centering
    \small
    \begin{subtable}[t]{.48\linewidth}
        \centering
        \begin{tabular}{cc}
        \toprule
        \textbf{Token Position} & \textbf{silh. score} \\
        \midrule
        \textit{first}        & .22 \\
        \textit{mid} & .39 \\
\rowcolor{gray!20}        \textit{last}         & \textbf{.61} \\
        \textit{weighted avg. of all tokens}         & \textbf{.64} \\
        \end{tabular}
        \caption{token position}
        \label{tab:activation_selection_subtable1}
    \end{subtable}%
    \hspace{.1cm} 
    \begin{subtable}[t]{.48\linewidth}
        \centering
        \begin{tabular}{cc}
        \toprule
        \textbf{Layer Position} & \textbf{silh. score} \\
        \midrule
        \textit{shallow (1/3 depth)}   & .51 \\
        \textit{mid (1/2 depth)}       & \textbf{.61} \\
        \textit{deep (2/3 depth)}      & \textbf{.64} \\
\rowcolor{gray!20}        \textit{last} & \textbf{.61} \\
        \end{tabular}
        \caption{layer position}
        \label{tab:activation_selection_subtable2}
    \end{subtable}
    \caption{\textbf{Effects of token and layer position to extract activations.} Later tokens and deeper layeres produce better Delta Activations, with the $2/3$ depth layer slightly surpassing the last layer.}
\end{table}
\textbf{Robustness to training settings.} Do differences in training settings have a greater impact on Delta Activations than the choice of finetuned domains? To evaluate this, we systematically perturbed the training process for models within our domain clusters. In our main setting, the model pool is organized into 5 distinct domain clusters, with 3 models in each cluster trained using identical settings. To test the impact of a specific training configuration—for instance, learning rate—the three models within each of the 5 domain clusters were trained using three different learning rates respectively (e.g., model 1 with $1e^{-4}$, model 2 with $4e^{-4}$, model 3 with $1e^{-5}$ within a single domain cluster). This process was independently repeated for variations in the number of training examples and the number of training epochs, where we vary number of training examples by 100, 1000, and 10000 and number of epochs by 1, 2, and 3.

Table~\ref{tab:setting_ablation} presents results which measure the clustering quality on domains when subjected to such training variations. indicates that varied training settings generally did not break domain-specific clustering. Changes to the amount of training data or the number of epochs had minimal effect on the quality of these clusters, which remained comparable to those formed under identical training settings.
The different-learning-rate setting yields a lower silhouette score, as expected, since learning rate significantly impacts training dynamics and tends to increase within-cluster variation. These observations confirm that Delta Activations effectively identify finetuning domains despite common variations in training procedures. On the other hand, these results also show the strength of Delta Activations in identifying the nuanced differences within each cluster.

\begin{table}[ht]
    \centering
    \small
    \begin{tabular}{crrrr}
    \toprule
    \textbf{Training Setting} & \textbf{LLaMA} & \textbf{Gemma} & \textbf{Qwen} & \textbf{Avg.} \\
    \midrule
    
    Different number of training examples                         & .66 & .51 & .68 & .62 \\
    Different learning rates           & .53 & .37 & .23 & .38  \\
    Different training epochs                             & .62 & .59 & .51 & .57 \\
    \rowcolor{gray!20}    Identical training settings      & .65 & .55 & .65 & .61 \\
    \end{tabular}
    \vspace{.7em}
    \caption{\textbf{Delta Activations' embeddings are robust to training hyperparameters.} Models trained in varying settings still form tight domain-specific clusters, comparable to those trained identically.}
    \label{tab:setting_ablation}
\end{table}

\textbf{Beyond domains: clustering by dataset properties.} In our setting in Section \ref{sec:model_embed}, each domain corresponds to a well-defined task with relatively uniform answer structure (e.g., multiple-choice). Here, we ablate that structure. We construct a new model pool by finetuning 3 models on each of five distinct subsets of \texttt{Tulu v2} \citep{ivison2023camels}: \textsc{CoT}, \textsc{GPT4-Alpaca}, \textsc{ShareGPT}, \textsc{CodeAlpaca}, and \textsc{Science}. Unlike domain datasets, these are less semantically disjoint. Instead, they differ in instruction format, conversational structure, and expected output style—ranging from open-ended dialogue to multi-step chain-of-thought reasoning, code snippets, and longform factual answers. Crucially, this makes the output distribution far more diverse and less predictable. Results are presented in Table \ref{tab:sft-robustness}. With no consistent answer template, output sentence embeddings fail to reflect model specialization for LLaMA and Qwen, whereas Delta Activations continue to achieve decent clustering quality.

\begin{table}[ht]
    \centering
    \small
    \begin{tabular}{crrrr}
    \toprule
    \textbf{Embedding Space} & \textbf{LLaMA} & \textbf{Gemma} & \textbf{Qwen} & \textbf{Avg.} \\
    \midrule
    output sentence  embeddings   & .06 & .23 & $-$.03 & .08 \\
\rowcolor{gray!20}    Delta Activations   & \textbf{.33} & \textbf{.41} & \textbf{.48} & \textbf{.41} \\
    \end{tabular}
    \vspace{.7em}
    \caption{\textbf{Clustering quality (silhouette score) across Tulu v2 instruction splits.} Delta Activations remain effective despite diverse output structures and blurred instruction boundaries.}
    \label{tab:sft-robustness}
\end{table}

\textbf{Beyond SFT: clustering by preference optimization.} Our experiments thus far focused on the setting of Supervised Finetuning which use the token-level cross entropy loss. Preference optimization techniques \citep{rafailov2023direct, meng2024simpo} maximize the likelihood that preferred responses are ranked higher. The different supervision signal of preference optimization affects the activation differently. We explore whether Delta Activations still yield reliable clustering for models in this case. To construct the model pool, we perform preference optimization on \textsc{LLaMA-3.1-8B-Instruct} using three disjoint 3000-example splits for each of three preference optimization datasets, namely UltraFeedback \citep{cui2024ultrafeedback}, HumanLLM \citep{ccalik2025enhancing}, and MetaMath-DPO \citep{pal2024smaug, yu2023metamath}. This experiment yields a silhouette score of \textbf{0.93}, which shows that Delta Activations can effectively capture similarity for preference optimization.

\begin{figure}[ht!]
    \centering
    \begin{minipage}[c]{0.4627\textwidth} 
        \centering
        \includegraphics[width=\linewidth]{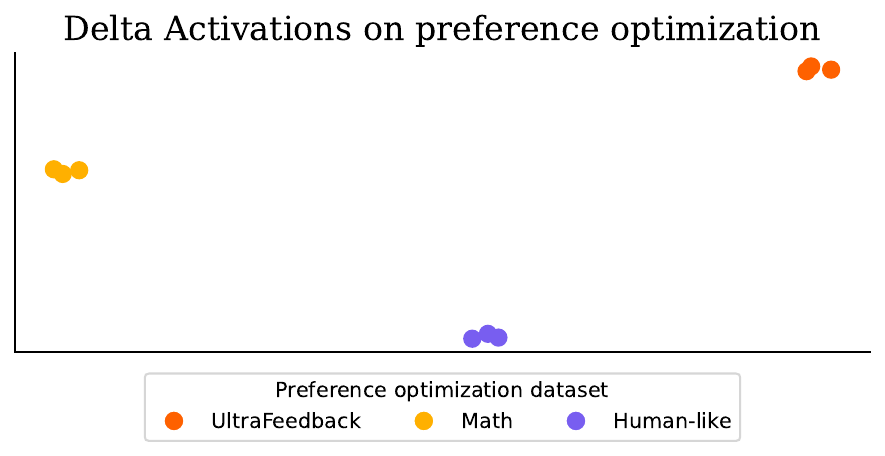}
        \caption{\textbf{Preference optimization.} Delta Activations can cluster models trained with DPO.} 
        \label{fig:tsne-dpo} 
    \end{minipage}%
    \hfill
    \begin{minipage}[c]{0.48\textwidth} 
        \centering
        \includegraphics[width=\linewidth]{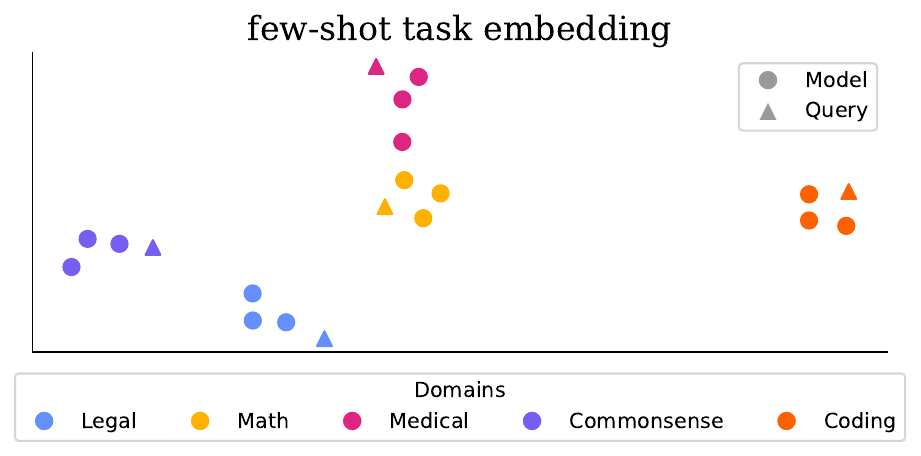}
        \caption{\textbf{Task embedding.} Few-shot task embedding is able to locate model clusters.} 
        \label{fig:few-shot} 
    \end{minipage}

\end{figure}
\subsection{Extensions}
\label{sec:task-embed}
In this section, we explore how Delta Activations as a model embedding can be extended to embedding a task, representing LLMs finetuned from different base LLMs, and guiding model merging.

\textbf{Task embedding.} We explore whether Delta Activations can embed tasks using only a few examples—a setting analogous to few-shot generalization. For each of the five domains in Section \ref{sec:model_embed}, we finetune the base LLM on 20 held-out examples that were not part of any previous training split. The detailed few-shot finetuning setting is in Appendix \ref{appendix:fewshot}. We then embed the few-shot finetuned models using Delta Activations, and use this embedding as a representation of the task.

We examine whether the task embedding can successfully locate the corresponding clusters on the three model pools constructed in Section \ref{sec:model_embed}. We define the \emph{retrieval rate} metric as the fraction of few-shot task embeddings that correctly retrieve their corresponding full-model cluster via nearest-neighbour search using cosine similarity. Gemma achieves 100\% retrieval rate while there is one failure case for each of LLaMA and Qwen. We present visualization of Gemma in Figure~\ref{fig:few-shot}, which shows that few-shot queries reliably align with their corresponding full model clusters (circles). Despite being trained on only 20 examples, the resulting Delta Activations recover the domain cluster. This suggests that Delta Activations on few-shot trained model can be an effective task embedding.

\begin{figure}[ht!]
    \centering
    \begin{subfigure}[c]{0.49\textwidth}
        \centering
        \includegraphics[width=\linewidth]{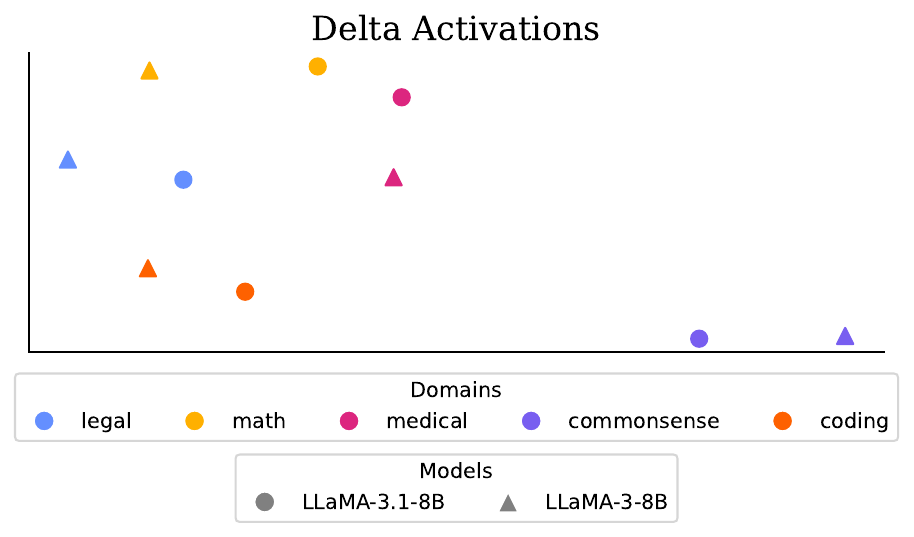}
        \caption{cross-checkpoint} 
        \label{fig:tsne-3130}
    \end{subfigure}%
    \hfill
    \begin{subfigure}[c]{0.49\textwidth}
        \centering
        \includegraphics[width=\linewidth]{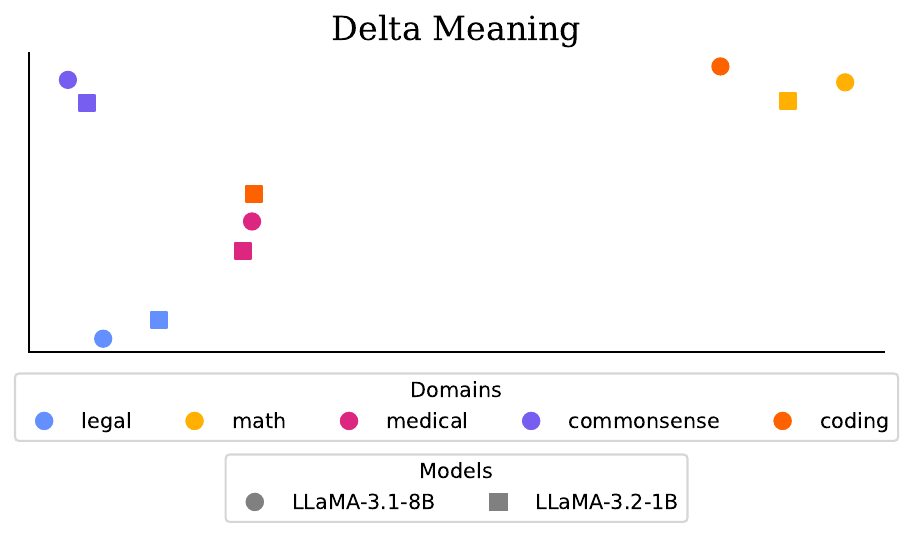}
        \caption{cross-architecture} 
        \label{fig:tsne-8b1b}
    \end{subfigure}
    
    \caption{\textbf{Cross-base-model clustering.} 
    (a) Delta Activations form domain clusters across models finetuned from \textsc{LLaMA-3.1-8B} and \textsc{LLaMA-3-8B}. 
    (b) Delta Meaning form domain clusters across finetuned models of different sizes, \textsc{LLaMA-3.1-8B} and \textsc{LLaMA-3.2-1B}.}
    \label{fig:cross-arch}
\end{figure}

\textbf{Cross-base-model clustering.}  
Delta Activations represent a finetuned model as the difference between its hidden states and those of its base model on the same inputs; therefore, they can only be directly applied to models derived from the \emph{same} base. Interestingly, we find that this delta signal can transfer across bases. To test this, we evaluate two settings: cross-checkpoint and cross-architecture. In a cross-checkpoint setting (\textsc{LLaMA-3-8B} vs. \textsc{LLaMA-3.1-8B}; 10 models over 5 domains), Delta Activations achieved a silhouette score of \textbf{0.39}, cleanly recovering the five domain-specialization clusters (Figure~\ref{fig:tsne-3130}). In a cross-architecture setting (\textsc{LLaMA-3.1-8B} vs. \textsc{LLaMA-3.2-1B}; 10 models), Delta Activations are no longer feasible because the embedding dimension differs across architectures, projecting models into vectors of incompatible sizes. Instead, we adopt \emph{Delta Meaning} (full implementation details in Appendix~\ref{appendix:delta-meaning}), which is architecture-agnostic, and it successfully forms four out of five domain clusters with a silhouette score of \textbf{0.32} (Figure~\ref{fig:tsne-8b1b}).

\textbf{Model selection.} LoraHub \citep{huang2023lorahub} hosts $\sim200$ finetuned models based on \textsc{FLAN-T5}. The method is evaluated on the Big-Bench Hard (BBH) \citep{suzgun2022challenging}. For each task in BBH, LoraHub randomly select 20 LoRAs and optimizes their merging coefficients over few-shot examples from the target task. Our approach can be easily applied to this scenario by embedding the corresponding task using the provided few-shot examples and leverage Delta Activations similarity to replace the random model selection. Specifically, we identify the single most-related LoRA model as an anchor and samples the remaining 19 models randomly. This simple selection strategy enabled by Delta Activations yields an average performance improvement of 2.0\% by boosting the average accuracy from 34.3\% to 36.3\% on the 26 tasks of the BBH benchmark. Interestingly, selecting 20 most similar models using Delta Activations yield an average performance of 30.3\%, which is far lower than random selection. We conjecture that this is due to model interference identified by \citet{NEURIPS2023_d28077e5}, where similar models are entangled with each other, resulting in bad merging performance. However, this in turn shows that Delta Activations are informative about the relationships between models. While model merging is beyond the scope of our work, Delta Activations enable model merging strategies beyond nearest-neighbour selection. Our proof-of-concept experiment focused on simple selection strategies, but Delta Activations could be leveraged for more sophisticated approaches. For instance, one could use Delta Activations to identify a maximally dispersed subset of models to mitigate interference or train a neural network on top of Delta Activations embeddings to optimize merging decisions.

\section{Related Work}
\textbf{Embedding Models and Tasks.}
\citet{ilharco2022editing} discovers that task vectors from finetuning can be merged to achieve multi-task learning. 
There are several attempts to represent LLMs using adapter weights~\citep{ostapenko2024towards, tan2025learnware}, evaluation profiles~\citep{zhuang2025embedllm}, or training-data-dependent characteristics~\citep{zhao-etal-2024-loraretriever}, but these methods rely on potentially inaccessible data or fail to reflect internal behavior. Our work differs by requiring no external supervision or evaluation, instead deriving embeddings from internal behaviors.

\textbf{Activations.}  
Recent work uncovers structure in hidden activations of LLMs, understanding how massive activation act as biases to steer LLM output \citep{sun2024massive,dettmers2022gpt3}. 
Activations are also central to post-training compression: they are used to compute weight saliency for pruning \citep{frantar2023sparsegpt,sun2024wanda,yin2023outlier} and to reduce quantization error using calibration sets \citep{frantar2022gptq,lin2023awq, zhang2023dyna}, which inspire the use of the probe dataset of Delta Activations. 
Additionally, learned activation shifts have been used to edit or transfer behavior across models \citep{liu2023context, li2023emergent, luo2024pace}. 

\textbf{Utilizing trained models.}
The landscape evolves from reusing a single model to multiple trained models. Finetuning \citep{zhuang2020comprehensive} is a common framework to build on top of a single pretrained model, which becomes the common practice for LLMs \citep{chung2024scaling, rafailov2023direct}. Other works use outputs \citep{hinton2015distilling, tian2019contrastive} or weights \citep{lecun1989optimal, han2015learning} of a pretrained model for the creation or initialization \citep{xu2024initializing, xia2024sheared} of more efficient models.
Efforts are made to effectively leverage multiple trained models through retrieval \citep{zhao-etal-2024-loraretriever, jin2024retrieval, luo2024stylus, kahana2025can}, composition \citep{yang2024model, chronopoulou2023adaptersoup,huang2023lorahub, feng2024model}, or routing \citep{lu2023routing, ong2024routellm, shnitzer2023large}. 

\textbf{Buiding model hubs.} Recent works \citep{horwitz2025unsupervised, yu2025neural} study the public model pool and can systematically uncover finetuning relationship among trained models. Lorahub \citep{huang2023lorahub}, LoraRetriever \citep{zhao-etal-2024-loraretriever}, and Learnware \citep{tan2025learnware} create model hubs involving from $\sim 40$ to $\sim 200$ models. It is also possible to create larger model hubs via neural network parameter diffusion \citep{wang2024neural, liang2025drag}, though achieving true diversity \citep{zeng2025generative} may require further development. 

\section{Discussion}
\label{sec:discussion}
Delta Activations provide a simple yet powerful way to represent finetuned LLMs by measuring shifts in their internal activations relative to a base LLM. Our experiments show that this representation consistently forms distinct clusters that reflect finetuning domains and offer the advantage of an additive property that mirrors multi-domain behavior. The stability of Delta Activations across varying finetuning settings shows its reliability for use-cases of model selection and merging in model hubs. We believe that Delta Activations can serve as a cornerstone for navigating the expanding landscape of finetuned models by enabling more efficient model discovery and reuse.

\textbf{Limitations and future work.} Delta Activations introduces a novel way to represent finetuned models but also poses practical considerations. Our experiments focused on three prominent open-source backbones but further evaluation on other architectures would be valuable to understand its broader applicability. In addition, Delta Activations require access to internal hidden states, which is not feasible to be evaluated on proprietary models. 

It is natural to ask how our method might perform on model pools substantially larger than those considered in our evaluation. Such a practical exercise would be most meaningful if the pool consists of models with diverse and uniquely valuable capabilities. While this is an interesting direction we intend to explore, we conjecture that today such pools exist primarily in proprietary settings (e.g. finetuned GPT models), and we hope our approach could facilitate sharing such models in future.

\newpage

\bibliography{references}

\begin{thebibliography}{79}
\providecommand{\natexlab}[1]{#1}
\providecommand{\url}[1]{\texttt{#1}}
\expandafter\ifx\csname urlstyle\endcsname\relax
  \providecommand{\doi}[1]{doi: #1}\else
  \providecommand{\doi}{doi: \begingroup \urlstyle{rm}\Url}\fi

\bibitem[Bai et~al.(2022)Bai, Jones, Ndousse, Askell, Chen, DasSarma, Drain, Fort, Ganguli, Henighan, et~al.]{bai2022training}
Yuntao Bai, Andy Jones, Kamal Ndousse, Amanda Askell, Anna Chen, Nova DasSarma, Dawn Drain, Stanislav Fort, Deep Ganguli, Tom Henighan, et~al.
\newblock Training a helpful and harmless assistant with reinforcement learning from human feedback.
\newblock \emph{arXiv preprint arXiv:2204.05862}, 2022.

\bibitem[Bolya et~al.(2025)Bolya, Huang, Sun, Cho, Madotto, Wei, Ma, Zhi, Rajasegaran, Rasheed, Wang, Monteiro, Xu, Dong, Ravi, Li, Doll{\'a}r, and Feichtenhofer]{bolya2025PerceptionEncoder}
Daniel Bolya, Po-Yao Huang, Peize Sun, Jang~Hyun Cho, Andrea Madotto, Chen Wei, Tengyu Ma, Jiale Zhi, Jathushan Rajasegaran, Hanoona Rasheed, Junke Wang, Marco Monteiro, Hu~Xu, Shiyu Dong, Nikhila Ravi, Daniel Li, Piotr Doll{\'a}r, and Christoph Feichtenhofer.
\newblock Perception encoder: The best visual embeddings are not at the output of the network.
\newblock \emph{arXiv:2504.13181}, 2025.

\bibitem[{\c{C}}al{\i}k \& Akku{\c{s}}(2025){\c{C}}al{\i}k and Akku{\c{s}}]{ccalik2025enhancing}
Ethem~Ya{\u{g}}{\i}z {\c{C}}al{\i}k and Talha~R{\"u}zgar Akku{\c{s}}.
\newblock Enhancing human-like responses in large language models.
\newblock \emph{arXiv preprint arXiv:2501.05032}, 2025.

\bibitem[Chen et~al.(2020)Chen, Radford, Child, Wu, Jun, Luan, and Sutskever]{chen2020generative}
Mark Chen, Alec Radford, Rewon Child, Jeffrey Wu, Heewoo Jun, David Luan, and Ilya Sutskever.
\newblock Generative pretraining from pixels.
\newblock In \emph{ICML}, 2020.

\bibitem[Chronopoulou et~al.(2023)Chronopoulou, Peters, Fraser, and Dodge]{chronopoulou2023adaptersoup}
Alexandra Chronopoulou, Matthew~E Peters, Alexander Fraser, and Jesse Dodge.
\newblock Adaptersoup: Weight averaging to improve generalization of pretrained language models.
\newblock In \emph{ACL}, 2023.

\bibitem[Chung et~al.(2024)Chung, Hou, Longpre, Zoph, Tay, Fedus, Li, Wang, Dehghani, Brahma, et~al.]{chung2024scaling}
Hyung~Won Chung, Le~Hou, Shayne Longpre, Barret Zoph, Yi~Tay, William Fedus, Yunxuan Li, Xuezhi Wang, Mostafa Dehghani, Siddhartha Brahma, et~al.
\newblock Scaling instruction-finetuned language models.
\newblock \emph{Journal of Machine Learning Research}, 2024.

\bibitem[Cobbe et~al.(2021)Cobbe, Kosaraju, Bavarian, Chen, Jun, Kaiser, Plappert, Tworek, Hilton, Nakano, Hesse, and Schulman]{cobbe2021gsm8k}
Karl Cobbe, Vineet Kosaraju, Mohammad Bavarian, Mark Chen, Heewoo Jun, Lukasz Kaiser, Matthias Plappert, Jerry Tworek, Jacob Hilton, Reiichiro Nakano, Christopher Hesse, and John Schulman.
\newblock Training verifiers to solve math word problems.
\newblock \emph{arXiv preprint arXiv:2110.14168}, 2021.

\bibitem[{CodeChef}(2009)]{codechef}
{CodeChef}.
\newblock Codechef.
\newblock \url{https://www.codechef.com}, 2009.
\newblock Competitive programming platform.

\bibitem[Cui et~al.(2024)Cui, Yuan, Ding, Yao, He, Zhu, Ni, Xie, Xie, Lin, et~al.]{cui2024ultrafeedback}
Ganqu Cui, Lifan Yuan, Ning Ding, Guanming Yao, Bingxiang He, Wei Zhu, Yuan Ni, Guotong Xie, Ruobing Xie, Yankai Lin, et~al.
\newblock Ultrafeedback: Boosting language models with scaled ai feedback.
\newblock In \emph{ICML}, 2024.

\bibitem[Dettmers et~al.(2022)Dettmers, Lewis, Belkada, and Zettlemoyer]{dettmers2022gpt3}
Tim Dettmers, Mike Lewis, Younes Belkada, and Luke Zettlemoyer.
\newblock Gpt3. int8 (): 8-bit matrix multiplication for transformers at scale.
\newblock \emph{Advances in neural information processing systems}, pp.\  30318--30332, 2022.

\bibitem[Feng et~al.(2024)Feng, Wang, Wang, Ebrahimi, Palangi, Miculicich, Kulshrestha, Rauschmayr, Choi, Tsvetkov, et~al.]{feng2024model}
Shangbin Feng, Zifeng Wang, Yike Wang, Sayna Ebrahimi, Hamid Palangi, Lesly Miculicich, Achin Kulshrestha, Nathalie Rauschmayr, Yejin Choi, Yulia Tsvetkov, et~al.
\newblock Model swarms: Collaborative search to adapt llm experts via swarm intelligence.
\newblock \emph{arXiv preprint arXiv:2410.11163}, 2024.

\bibitem[Frantar \& Alistarh(2023)Frantar and Alistarh]{frantar2023sparsegpt}
Elias Frantar and Dan Alistarh.
\newblock Sparsegpt: Massive language models can be accurately pruned in one-shot.
\newblock In \emph{International Conference on Machine Learning}, pp.\  10323--10337, 2023.

\bibitem[Frantar et~al.(2022)Frantar, Ashkboos, Hoefler, and Alistarh]{frantar2022gptq}
Elias Frantar, Saleh Ashkboos, Torsten Hoefler, and Dan Alistarh.
\newblock Gptq: Accurate post-training quantization for generative pre-trained transformers.
\newblock \emph{arXiv preprint arXiv:2210.17323}, 2022.

\bibitem[{FreedomIntelligence}(2024)]{freedomintelligence2024disease}
{FreedomIntelligence}.
\newblock Disease database.
\newblock \url{https://huggingface.co/datasets/FreedomIntelligence/Disease_Database}, 2024.
\newblock Hugging Face dataset.

\bibitem[Guha et~al.(2023)Guha, Nyarko, Ho, Ré, Chilton, Narayana, Chohlas-Wood, Peters, Waldon, Rockmore, Zambrano, Talisman, Hoque, Surani, Fagan, Sarfaty, Dickinson, Porat, Hegland, Wu, Nudell, Niklaus, Nay, Choi, Tobia, Hagan, Ma, Livermore, Rasumov-Rahe, Holzenberger, Kolt, Henderson, Rehaag, Goel, Gao, Williams, Gandhi, Zur, Iyer, and Li]{guha2023legalbench}
Neel Guha, Julian Nyarko, Daniel~E. Ho, Christopher Ré, Adam Chilton, Aditya Narayana, Alex Chohlas-Wood, Austin Peters, Brandon Waldon, Daniel~N. Rockmore, Diego Zambrano, Dmitry Talisman, Enam Hoque, Faiz Surani, Frank Fagan, Galit Sarfaty, Gregory~M. Dickinson, Haggai Porat, Jason Hegland, Jessica Wu, Joe Nudell, Joel Niklaus, John Nay, Jonathan~H. Choi, Kevin Tobia, Margaret Hagan, Megan Ma, Michael Livermore, Nikon Rasumov-Rahe, Nils Holzenberger, Noam Kolt, Peter Henderson, Sean Rehaag, Sharad Goel, Shang Gao, Spencer Williams, Sunny Gandhi, Tom Zur, Varun Iyer, and Zehua Li.
\newblock Legalbench: A collaboratively built benchmark for measuring legal reasoning in large language models, 2023.

\bibitem[Han et~al.(2015)Han, Pool, Tran, and Dally]{han2015learning}
Song Han, Jeff Pool, John Tran, and William Dally.
\newblock Learning both weights and connections for efficient neural network.
\newblock In \emph{NeurIPS}, 2015.

\bibitem[He et~al.(2024)He, Hu, Lin, Zhang, and Zhao]{he2024localize_and_stitch}
Yifei He, Yuzheng Hu, Yong Lin, Tong Zhang, and Han Zhao.
\newblock Localize-and-stitch: Efficient model merging via sparse task arithmetic.
\newblock \emph{Transactions on Machine Learning Research}, 2024, 2024.
\newblock Preprint available at arXiv:2408.13656.

\bibitem[Hinton et~al.(2015)Hinton, Vinyals, and Dean]{hinton2015distilling}
Geoffrey Hinton, Oriol Vinyals, and Jeff Dean.
\newblock Distilling the knowledge in a neural network.
\newblock \emph{arXiv preprint arXiv:1503.02531}, 2015.

\bibitem[Horwitz et~al.(2025)Horwitz, Shul, and Hoshen]{horwitz2025unsupervised}
Eliahu Horwitz, Asaf Shul, and Yedid Hoshen.
\newblock Unsupervised model tree heritage recovery.
\newblock In \emph{ICLR}, 2025.

\bibitem[Hu et~al.(2022)Hu, Shen, Wallis, Allen-Zhu, Li, Wang, Wang, and Chen]{hu2022lora}
Edward~J Hu, Yelong Shen, Phillip Wallis, Zeyuan Allen-Zhu, Yuanzhi Li, Shean Wang, Lu~Wang, and Weizhu Chen.
\newblock Lo{RA}: Low-rank adaptation of large language models.
\newblock In \emph{ICLR}, 2022.

\bibitem[Huang et~al.(2024{\natexlab{a}})Huang, Liu, Lin, Pang, Du, and Lin]{huang2023lorahub}
Chengsong Huang, Qian Liu, Bill~Yuchen Lin, Tianyu Pang, Chao Du, and Min Lin.
\newblock Lorahub: Efficient cross-task generalization via dynamic lora composition.
\newblock In \emph{COLM}, 2024{\natexlab{a}}.

\bibitem[Huang et~al.(2024{\natexlab{b}})Huang, Cheng, Liu, Hao, Song, Xu, Yang, Liu, Zhang, Chai, et~al.]{Huang2024OpenCoderTO}
Siming Huang, Tianhao Cheng, Jason~Klein Liu, Jiaran Hao, Liuyihan Song, Yang Xu, J~Yang, JH~Liu, Chenchen Zhang, Linzheng Chai, et~al.
\newblock Opencoder: The open cookbook for top-tier code large language models.
\newblock \emph{arXiv preprint arXiv:2411.04905}, 2024{\natexlab{b}}.

\bibitem[Ilharco et~al.(2023)Ilharco, Ribeiro, Wortsman, Gururangan, Schmidt, Hajishirzi, and Farhadi]{ilharco2022editing}
Gabriel Ilharco, Marco~Tulio Ribeiro, Mitchell Wortsman, Suchin Gururangan, Ludwig Schmidt, Hannaneh Hajishirzi, and Ali Farhadi.
\newblock Editing models with task arithmetic.
\newblock In \emph{ICLR}, 2023.

\bibitem[Ivison et~al.(2023)Ivison, Wang, Pyatkin, Lambert, Peters, Dasigi, Jang, Wadden, Smith, Beltagy, and Hajishirzi]{ivison2023camels}
Hamish Ivison, Yizhong Wang, Valentina Pyatkin, Nathan Lambert, Matthew Peters, Pradeep Dasigi, Joel Jang, David Wadden, Noah~A. Smith, Iz~Beltagy, and Hannaneh Hajishirzi.
\newblock Camels in a changing climate: Enhancing lm adaptation with tulu 2, 2023.

\bibitem[Jin et~al.(2020)Jin, Pan, Oufattole, Weng, Fang, and Szolovits]{jin2020disease}
Di~Jin, Eileen Pan, Nassim Oufattole, Wei-Hung Weng, Hanyi Fang, and Peter Szolovits.
\newblock What disease does this patient have? a large-scale open domain question answering dataset from medical exams.
\newblock \emph{arXiv preprint arXiv:2009.13081}, 2020.

\bibitem[Jin et~al.(2024)Jin, Shu, Kim, Xiao, Song, Chen, Liu, Li, and Li]{jin2024retrieval}
Pengfei Jin, Peng Shu, Sekeun Kim, Qing Xiao, Sifan Song, Cheng Chen, Tianming Liu, Xiang Li, and Quanzheng Li.
\newblock Retrieval instead of fine-tuning: A retrieval-based parameter ensemble for zero-shot learning.
\newblock \emph{arXiv preprint arXiv:2410.09908}, 2024.

\bibitem[Jin et~al.(2019)Jin, Dhingra, Liu, Cohen, and Lu]{jin2019pubmedqa}
Qiao Jin, Bhuwan Dhingra, Zhengping Liu, William Cohen, and Xinghua Lu.
\newblock Pubmedqa: A dataset for biomedical research question answering.
\newblock In \emph{EMNLP-IJCNLP}, 2019.

\bibitem[Kahana et~al.(2025)Kahana, Nathan, Horwitz, and Hoshen]{kahana2025can}
Jonathan Kahana, Or~Nathan, Eliahu Horwitz, and Yedid Hoshen.
\newblock Can this model also recognize dogs? zero-shot model search from weights.
\newblock \emph{arXiv preprint arXiv:2502.09619}, 2025.

\bibitem[Koren et~al.(2009)Koren, Bell, and Volinsky]{koren2009matrix}
Yehuda Koren, Robert Bell, and Chris Volinsky.
\newblock Matrix factorization techniques for recommender systems.
\newblock \emph{Computer}, 2009.

\bibitem[Krizhevsky et~al.(2012)Krizhevsky, Sutskever, and Hinton]{krizhevsky2012imagenet}
Alex Krizhevsky, Ilya Sutskever, and Geoffrey~E Hinton.
\newblock Imagenet classification with deep convolutional neural networks.
\newblock In \emph{NIPS}, 2012.

\bibitem[LeCun et~al.(1989)LeCun, Denker, and Solla]{lecun1989optimal}
Yann LeCun, John Denker, and Sara Solla.
\newblock Optimal brain damage.
\newblock In \emph{NeurIPS}, 1989.

\bibitem[Li et~al.(2023)Li, Hopkins, Bau, Vi{\'e}gas, Pfister, and Wattenberg]{li2023emergent}
Kenneth Li, Aspen~K Hopkins, David Bau, Fernanda Vi{\'e}gas, Hanspeter Pfister, and Martin Wattenberg.
\newblock Emergent world representations: Exploring a sequence model trained on a synthetic task.
\newblock \emph{ICLR}, 2023.

\bibitem[Liang et~al.(2025)Liang, Tang, Zhou, Zhao, Shi, Zhao, Li, Wang, Sch{\"u}rholt, Borth, et~al.]{liang2025drag}
Zhiyuan Liang, Dongwen Tang, Yuhao Zhou, Xuanlei Zhao, Mingjia Shi, Wangbo Zhao, Zekai Li, Peihao Wang, Konstantin Sch{\"u}rholt, Damian Borth, et~al.
\newblock Drag-and-drop llms: Zero-shot prompt-to-weights.
\newblock \emph{arXiv preprint arXiv:2506.16406}, 2025.

\bibitem[Lin et~al.(2023)Lin, Tang, Tang, Yang, Dang, and Han]{lin2023awq}
Ji~Lin, Jiaming Tang, Haotian Tang, Shang Yang, Xingyu Dang, and Song Han.
\newblock Awq: Activation-aware weight quantization for llm compression and acceleration.
\newblock \emph{MlSys}, 2023.

\bibitem[Liu et~al.(2024{\natexlab{a}})Liu, Feng, Xue, Wang, Wu, Lu, Zhao, Deng, Zhang, Ruan, et~al.]{liu2024deepseek}
Aixin Liu, Bei Feng, Bing Xue, Bingxuan Wang, Bochao Wu, Chengda Lu, Chenggang Zhao, Chengqi Deng, Chenyu Zhang, Chong Ruan, et~al.
\newblock Deepseek-v3 technical report.
\newblock \emph{arXiv preprint arXiv:2412.19437}, 2024{\natexlab{a}}.

\bibitem[Liu et~al.(2023)Liu, Ye, Xing, and Zou]{liu2023context}
Sheng Liu, Haotian Ye, Lei Xing, and James Zou.
\newblock In-context vectors: Making in context learning more effective and controllable through latent space steering.
\newblock \emph{arXiv preprint arXiv:2311.06668}, 2023.

\bibitem[Liu et~al.(2024{\natexlab{b}})Liu, Trager, Achille, Perera, Zancato, and Soatto]{liu2024meaning}
Tian~Yu Liu, Matthew Trager, Alessandro Achille, Pramuditha Perera, Luca Zancato, and Stefano Soatto.
\newblock Meaning representations from trajectories in autoregressive models.
\newblock In \emph{ICLR}, 2024{\natexlab{b}}.

\bibitem[Lu et~al.(2023)Lu, Yuan, Lin, Lin, Yuan, Zhou, and Zhou]{lu2023routing}
Keming Lu, Hongyi Yuan, Runji Lin, Junyang Lin, Zheng Yuan, Chang Zhou, and Jingren Zhou.
\newblock Routing to the expert: Efficient reward-guided ensemble of large language models.
\newblock \emph{arXiv preprint arXiv:2311.08692}, 2023.

\bibitem[Luo et~al.(2024{\natexlab{a}})Luo, Ding, Chan, Thaker, Chattopadhyay, Callison-Burch, and Vidal]{luo2024pace}
Jinqi Luo, Tianjiao Ding, Kwan Ho~Ryan Chan, Darshan Thaker, Aditya Chattopadhyay, Chris Callison-Burch, and Ren{\'e} Vidal.
\newblock Pace: Parsimonious concept engineering for large language models.
\newblock In \emph{NeurIPS}, 2024{\natexlab{a}}.

\bibitem[Luo et~al.(2024{\natexlab{b}})Luo, Wong, Trabucco, Huang, Gonzalez, Salakhutdinov, Stoica, et~al.]{luo2024stylus}
Michael Luo, Justin Wong, Brandon Trabucco, Yanping Huang, Joseph~E Gonzalez, Ruslan Salakhutdinov, Ion Stoica, et~al.
\newblock Stylus: Automatic adapter selection for diffusion models.
\newblock In \emph{NeurIPS}, 2024{\natexlab{b}}.

\bibitem[Luo et~al.(2024{\natexlab{c}})Luo, Xu, Zhao, Sun, Geng, Hu, Tao, Ma, Lin, and Jiang]{luo2023wizardcoder}
Ziyang Luo, Can Xu, Pu~Zhao, Qingfeng Sun, Xiubo Geng, Wenxiang Hu, Chongyang Tao, Jing Ma, Qingwei Lin, and Daxin Jiang.
\newblock Wizardcoder: Empowering code large language models with evol-instruct.
\newblock \emph{ICLR}, 2024{\natexlab{c}}.

\bibitem[Meng et~al.(2024)Meng, Xia, and Chen]{meng2024simpo}
Yu~Meng, Mengzhou Xia, and Danqi Chen.
\newblock Simpo: Simple preference optimization with a reference-free reward.
\newblock In \emph{NeurIPS}, 2024.

\bibitem[Mikolov et~al.(2013)Mikolov, Chen, Corrado, and Dean]{mikolov2013efficient}
Tomas Mikolov, Kai Chen, Greg Corrado, and Jeffrey Dean.
\newblock Efficient estimation of word representations in vector space.
\newblock In \emph{ICLR Workshop}, 2013.

\bibitem[{Mohamed-Ahmed161}(2024)]{mohamedahmed2024diseasesymptoms}
{Mohamed-Ahmed161}.
\newblock Disease-symptoms dataset.
\newblock \url{https://huggingface.co/datasets/Mohamed-Ahmed161/Disease-Symptoms}, 2024.
\newblock Hugging Face dataset.

\bibitem[Muennighoff(2022)]{muennighoff2022sgpt}
Niklas Muennighoff.
\newblock Sgpt: Gpt sentence embeddings for semantic search.
\newblock \emph{arXiv preprint arXiv:2202.08904}, 2022.

\bibitem[Ong et~al.(2024)Ong, Almahairi, Wu, Chiang, Wu, Gonzalez, Kadous, and Stoica]{ong2024routellm}
Isaac Ong, Amjad Almahairi, Vincent Wu, Wei-Lin Chiang, Tianhao Wu, Joseph~E Gonzalez, M~Waleed Kadous, and Ion Stoica.
\newblock Routellm: Learning to route llms from preference data.
\newblock In \emph{ICLR}, 2024.

\bibitem[Ortiz-Jimenez et~al.(2023)Ortiz-Jimenez, Favero, and Frossard]{NEURIPS2023_d28077e5}
Guillermo Ortiz-Jimenez, Alessandro Favero, and Pascal Frossard.
\newblock Task arithmetic in the tangent space: Improved editing of pre-trained models.
\newblock In A.~Oh, T.~Naumann, A.~Globerson, K.~Saenko, M.~Hardt, and S.~Levine (eds.), \emph{NeurIPS}, 2023.

\bibitem[Ostapenko et~al.(2024)Ostapenko, Su, Ponti, Charlin, Roux, Pereira, Caccia, and Sordoni]{ostapenko2024towards}
Oleksiy Ostapenko, Zhan Su, Edoardo~Maria Ponti, Laurent Charlin, Nicolas~Le Roux, Matheus Pereira, Lucas Caccia, and Alessandro Sordoni.
\newblock Towards modular llms by building and reusing a library of loras.
\newblock \emph{arXiv preprint arXiv:2405.11157}, 2024.

\bibitem[Pal et~al.(2022)Pal, Umapathi, and Sankarasubbu]{pmlr-v174-pal22a}
Ankit Pal, Logesh~Kumar Umapathi, and Malaikannan Sankarasubbu.
\newblock Medmcqa: A large-scale multi-subject multi-choice dataset for medical domain question answering.
\newblock In \emph{Proceedings of the Conference on Health, Inference, and Learning}, 2022.

\bibitem[Pal et~al.(2024)Pal, Karkhanis, Dooley, Roberts, Naidu, and White]{pal2024smaug}
Arka Pal, Deep Karkhanis, Samuel Dooley, Manley Roberts, Siddartha Naidu, and Colin White.
\newblock Smaug: Fixing failure modes of preference optimisation with dpo-positive.
\newblock \emph{arXiv preprint arXiv:2402.13228}, 2024.

\bibitem[Quan et~al.(2025)Quan, Yang, Yu, Zheng, Liu, Yang, Ren, Gao, Miao, Feng, Wang, Yang, Cui, Fan, Zhang, Hui, and Lin]{quan2025codeelobenchmarkingcompetitionlevelcode}
Shanghaoran Quan, Jiaxi Yang, Bowen Yu, Bo~Zheng, Dayiheng Liu, An~Yang, Xuancheng Ren, Bofei Gao, Yibo Miao, Yunlong Feng, Zekun Wang, Jian Yang, Zeyu Cui, Yang Fan, Yichang Zhang, Binyuan Hui, and Junyang Lin.
\newblock Codeelo: Benchmarking competition-level code generation of llms with human-comparable elo ratings, 2025.

\bibitem[Rafailov et~al.(2023)Rafailov, Sharma, Mitchell, Manning, Ermon, and Finn]{rafailov2023direct}
Rafael Rafailov, Archit Sharma, Eric Mitchell, Christopher~D Manning, Stefano Ermon, and Chelsea Finn.
\newblock Direct preference optimization: Your language model is secretly a reward model.
\newblock In \emph{NeurIPS}, 2023.

\bibitem[Ren \& Sutherland(2025)Ren and Sutherland]{ren2024learning}
Yi~Ren and Danica~J Sutherland.
\newblock Learning dynamics of llm finetuning.
\newblock In \emph{ICLR}, 2025.

\bibitem[Rousseeuw(1987)]{silhouettes}
Peter~J. Rousseeuw.
\newblock Silhouettes: A graphical aid to the interpretation and validation of cluster analysis.
\newblock \emph{Journal of Computational and Applied Mathematics}, 1987.

\bibitem[Shnitzer et~al.(2023)Shnitzer, Ou, Silva, Soule, Sun, Solomon, Thompson, and Yurochkin]{shnitzer2023large}
Tal Shnitzer, Anthony Ou, M{\'\i}rian Silva, Kate Soule, Yuekai Sun, Justin Solomon, Neil Thompson, and Mikhail Yurochkin.
\newblock Large language model routing with benchmark datasets.
\newblock \emph{arXiv preprint arXiv:2309.15789}, 2023.

\bibitem[Sun et~al.(2023)Sun, Liu, Bair, and Kolter]{sun2024wanda}
Mingjie Sun, Zhuang Liu, Anna Bair, and J~Zico Kolter.
\newblock A simple and effective pruning approach for large language models.
\newblock \emph{arXiv preprint arXiv:2306.11695}, 2023.

\bibitem[Sun et~al.(2024)Sun, Chen, Kolter, and Liu]{sun2024massive}
Mingjie Sun, Xinlei Chen, J.~Zico Kolter, and Zhuang Liu.
\newblock Massive activations in large language models.
\newblock \emph{COLM}, 2024.

\bibitem[Sun et~al.(2025)Sun, Yin, Xu, Kolter, and Liu]{sun2025idiosyncrasies}
Mingjie Sun, Yida Yin, Zhiqiu Xu, J.~Zico Kolter, and Zhuang Liu.
\newblock Idiosyncrasies in large language models.
\newblock In \emph{ICML}, 2025.

\bibitem[Suzgun et~al.(2022)Suzgun, Scales, Sch{\"a}rli, Gehrmann, Tay, Chung, Chowdhery, Le, Chi, Zhou, et~al.]{suzgun2022challenging}
Mirac Suzgun, Nathan Scales, Nathanael Sch{\"a}rli, Sebastian Gehrmann, Yi~Tay, Hyung~Won Chung, Aakanksha Chowdhery, Quoc~V Le, Ed~H Chi, Denny Zhou, et~al.
\newblock Challenging big-bench tasks and whether chain-of-thought can solve them.
\newblock \emph{arXiv preprint arXiv:2210.09261}, 2022.

\bibitem[Tan et~al.(2025)Tan, Zhao, Shi, Zhang, Tan, Yu, and Zhou]{tan2025learnware}
Zhi-Hao Tan, Zi-Chen Zhao, Hao-Yu Shi, Xin-Yu Zhang, Peng Tan, Yang Yu, and Zhi-Hua Zhou.
\newblock Learnware of language models: Specialized small language models can do big.
\newblock \emph{arXiv preprint arXiv:2505.13425}, 2025.

\bibitem[Taori et~al.(2023)Taori, Gulrajani, Zhang, Dubois, Li, Guestrin, Liang, and Hashimoto]{taori2023alpaca}
Rohan Taori, Ishaan Gulrajani, Tianyi Zhang, Yann Dubois, Xuechen Li, Carlos Guestrin, Percy Liang, and Tatsunori~B Hashimoto.
\newblock Alpaca: A strong, replicable instruction-following model.
\newblock \emph{Stanford Center for Research on Foundation Models}, 2023.

\bibitem[Team et~al.(2024)Team, Riviere, Pathak, Sessa, Hardin, Bhupatiraju, Hussenot, Mesnard, Shahriari, Ram{\'e}, et~al.]{team2024gemma}
Gemma Team, Morgane Riviere, Shreya Pathak, Pier~Giuseppe Sessa, Cassidy Hardin, Surya Bhupatiraju, L{\'e}onard Hussenot, Thomas Mesnard, Bobak Shahriari, Alexandre Ram{\'e}, et~al.
\newblock Gemma 2: Improving open language models at a practical size.
\newblock \emph{arXiv preprint arXiv:2408.00118}, 2024.

\bibitem[Tian et~al.(2019)Tian, Krishnan, and Isola]{tian2019contrastive}
Yonglong Tian, Dilip Krishnan, and Phillip Isola.
\newblock Contrastive representation distillation.
\newblock \emph{arXiv preprint arXiv:1910.10699}, 2019.

\bibitem[Touvron et~al.(2023)]{touvron2023llama}
Hugo Touvron et~al.
\newblock Llama: Open and efficient foundation language models.
\newblock \emph{arXiv preprint arXiv:2302.13971}, 2023.

\bibitem[Wang et~al.(2024)Wang, Tang, Zeng, Yin, Xu, Zhou, Zang, Darrell, Liu, and You]{wang2024neural}
Kai Wang, Dongwen Tang, Boya Zeng, Yida Yin, Zhaopan Xu, Yukun Zhou, Zelin Zang, Trevor Darrell, Zhuang Liu, and Yang You.
\newblock Neural network diffusion.
\newblock \emph{arXiv preprint arXiv:2402.13144}, 2024.

\bibitem[Wang et~al.(2021)Wang, Bao, Huang, Dong, and Wei]{minilmv2}
Wenhui Wang, Hangbo Bao, Shaohan Huang, Li~Dong, and Furu Wei.
\newblock {M}ini{LM}v2: Multi-head self-attention relation distillation for compressing pretrained transformers.
\newblock In \emph{ACL}, 2021.

\bibitem[Xia et~al.(2024)Xia, Gao, Zeng, and Chen]{xia2024sheared}
Mengzhou Xia, Tianyu Gao, Zhiyuan Zeng, and Danqi Chen.
\newblock Sheared {LL}a{MA}: Accelerating language model pre-training via structured pruning.
\newblock In \emph{ICLR}, 2024.

\bibitem[Xu et~al.(2024)Xu, Chen, Vishniakov, Yin, Shen, Darrell, Liu, and Liu]{xu2024initializing}
Zhiqiu Xu, Yanjie Chen, Kirill Vishniakov, Yida Yin, Zhiqiang Shen, Trevor Darrell, Lingjie Liu, and Zhuang Liu.
\newblock Initializing models with larger ones.
\newblock In \emph{ICLR}, 2024.

\bibitem[Yang et~al.(2024{\natexlab{a}})Yang, Yang, Zhang, Hui, Zheng, Yu, Li, Liu, Huang, Wei, Lin, Yang, Tu, Zhang, Yang, Yang, Zhou, Lin, Dang, Lu, Bao, Yang, Yu, Li, Xue, Zhang, Zhu, Men, Lin, Li, Xia, Ren, Ren, Fan, Su, Zhang, Wan, Liu, Cui, Zhang, and Qiu]{qwen2.5}
An~Yang, Baosong Yang, Beichen Zhang, Binyuan Hui, Bo~Zheng, Bowen Yu, Chengyuan Li, Dayiheng Liu, Fei Huang, Haoran Wei, Huan Lin, Jian Yang, Jianhong Tu, Jianwei Zhang, Jianxin Yang, Jiaxi Yang, Jingren Zhou, Junyang Lin, Kai Dang, Keming Lu, Keqin Bao, Kexin Yang, Le~Yu, Mei Li, Mingfeng Xue, Pei Zhang, Qin Zhu, Rui Men, Runji Lin, Tianhao Li, Tingyu Xia, Xingzhang Ren, Xuancheng Ren, Yang Fan, Yang Su, Yichang Zhang, Yu~Wan, Yuqiong Liu, Zeyu Cui, Zhenru Zhang, and Zihan Qiu.
\newblock Qwen2.5 technical report.
\newblock \emph{arXiv preprint arXiv:2412.15115}, 2024{\natexlab{a}}.

\bibitem[Yang et~al.(2024{\natexlab{b}})Yang, Shen, Guo, Wang, Cao, Zhang, and Tao]{yang2024model}
Enneng Yang, Li~Shen, Guibing Guo, Xingwei Wang, Xiaochun Cao, Jie Zhang, and Dacheng Tao.
\newblock Model merging in llms, mllms, and beyond: Methods, theories, applications and opportunities.
\newblock \emph{arXiv preprint arXiv:2408.07666}, 2024{\natexlab{b}}.

\bibitem[Yin et~al.(2023)Yin, Wu, Zhang, Hsieh, Wang, Jia, Pechenizkiy, Liang, Wang, and Liu]{yin2023outlier}
Lu~Yin, You Wu, Zhenyu Zhang, Cheng-Yu Hsieh, Yaqing Wang, Yiling Jia, Mykola Pechenizkiy, Yi~Liang, Zhangyang Wang, and Shiwei Liu.
\newblock Outlier weighed layerwise sparsity (owl): A missing secret sauce for pruning llms to high sparsity.
\newblock \emph{arXiv preprint arXiv:2310.05175}, 2023.

\bibitem[Yu et~al.(2023)Yu, Jiang, Shi, Yu, Liu, Zhang, Kwok, Li, Weller, and Liu]{yu2023metamath}
Longhui Yu, Weisen Jiang, Han Shi, Jincheng Yu, Zhengying Liu, Yu~Zhang, James~T Kwok, Zhenguo Li, Adrian Weller, and Weiyang Liu.
\newblock Metamath: Bootstrap your own mathematical questions for large language models.
\newblock \emph{arXiv preprint arXiv:2309.12284}, 2023.

\bibitem[Yu \& Wang(2025)Yu and Wang]{yu2025neural}
Runpeng Yu and Xinchao Wang.
\newblock Neural phylogeny: Fine-tuning relationship detection among neural networks.
\newblock In \emph{ICLR}, 2025.

\bibitem[Zellers et~al.(2019)Zellers, Holtzman, Bisk, Farhadi, and Choi]{zellers2019hellaswag}
Rowan Zellers, Ari Holtzman, Yonatan Bisk, Ali Farhadi, and Yejin Choi.
\newblock Hellaswag: Can a machine really finish your sentence?
\newblock In \emph{ACL}, 2019.

\bibitem[Zeng et~al.(2025)Zeng, Yin, Xu, and Liu]{zeng2025generative}
Boya Zeng, Yida Yin, Zhiqiu Xu, and Zhuang Liu.
\newblock Generative modeling of weights: Generalization or memorization?
\newblock \emph{arXiv preprint arXiv:2506.07998}, 2025.

\bibitem[Zhang et~al.(2023)Zhang, Zhao, Lin, Sun, Yao, Han, Tanner, Liu, and Ji]{zhang2023dyna}
Yuxin Zhang, Lirui Zhao, Mingbao Lin, Yunyun Sun, Yiwu Yao, Xingjia Han, Jared Tanner, Shiwei Liu, and Rongrong Ji.
\newblock Dynamic sparse no training: Training-free fine-tuning for sparse llms.
\newblock \emph{arXiv preprint arXiv:2310.08915}, 2023.

\bibitem[Zhao et~al.(2024)Zhao, Gan, Wang, Zhou, Yang, Kuang, and Wu]{zhao-etal-2024-loraretriever}
Ziyu Zhao, Leilei Gan, Guoyin Wang, Wangchunshu Zhou, Hongxia Yang, Kun Kuang, and Fei Wu.
\newblock {L}ora{R}etriever: Input-aware {L}o{RA} retrieval and composition for mixed tasks in the wild.
\newblock In \emph{ACL Findings}, 2024.

\bibitem[Zhuang et~al.(2020)Zhuang, Qi, Duan, Xi, Zhu, Zhu, Xiong, and He]{zhuang2020comprehensive}
Fuzhen Zhuang, Zhiyuan Qi, Keyu Duan, Dongbo Xi, Yongchun Zhu, Hengshu Zhu, Hui Xiong, and Qing He.
\newblock A comprehensive survey on transfer learning.
\newblock \emph{Proceedings of the IEEE}, 2020.

\bibitem[Zhuang et~al.(2025)Zhuang, Wu, Wen, Li, Jiao, and Ramchandran]{zhuang2025embedllm}
Richard Zhuang, Tianhao Wu, Zhaojin Wen, Andrew Li, Jiantao Jiao, and Kannan Ramchandran.
\newblock Embed{LLM}: Learning compact representations of large language models.
\newblock In \emph{ICLR}, 2025.

\end{thebibliography}
\bibliographystyle{neurips_2025}


\newpage
\appendix

\newpage

\section{Training settings}
\label{appendix:trainingsett}
For LoRA, we set rank $r=8$, $\alpha=16$, targeting query, key, value, and MLP projections ({\tt q\_proj}, {\tt k\_proj}, {\tt v\_proj}, {\tt up\_proj}, {\tt down\_proj}, {\tt gate\_proj}), with no dropout and no bias parameters.

\subsection{Datasets}
\begin{table}[h]
\small
\centering
\begin{tabular}{ll}
\hline
\textbf{Dataset} & \textbf{Description} \\
\hline
OpenCoder-LLM & Educational programming instructions (opc-sft-stage2) \\
GSM8K & Grade-school math problems with chain-of-thought reasoning \\
HellaSwag & Commonsense natural language inference \\
LegalBench & Privacy policy question answering \\
PubMedQA & Medical Q\&A from PubMed abstracts \\
\hline
\end{tabular}
\vspace{0.7em}
\caption{\textbf{Summary of datasets used in the experiments.} Each dataset was split into three subsets of 3,000 examples.}
\label{tab:datasets}
\end{table}

\subsection{Prompt Templates}
\begin{table}[h]
\small
\setlength{\tabcolsep}{5pt}  
\centering
\begin{tabularx}{\linewidth}{>{\raggedright\arraybackslash}p{0.32\linewidth} X}
\toprule
\textbf{Prompt Type} & \textbf{Template (truncated)} \\ 
\midrule
Task-specific (Programming)  & \texttt{Below is an instruction that describes a programming task\ldots} \\
Task-specific (Math)         & \texttt{Below is a grade-school math problem. Please work through the reasoning step-by-step\ldots} \\
Task-specific (Legal)        & \texttt{Below is a legal-reasoning task from the LegalBench benchmark\ldots} \\
Task-specific (Medical)      & \texttt{Below is a medical question based on a PubMed article\ldots} \\
Task-specific (Commonsense)  & \texttt{Below is a scenario. What happens next in this paragraph\ldots } \\ 
Universal                    & \texttt{Below is an instruction that describes a task. Write a response that appropriately completes the request.} \\
\bottomrule
\end{tabularx}
\vspace{0.6em}
\caption{\textbf{Prompt templates used during training.} Task-specific templates are customized to each domain; the universal prompt is applied uniformly across all tasks during finetuning.}
\label{tab:prompts}
\end{table}

\subsection{Training Hyperparameters}
\begin{table}[h]
\small
\centering
\begin{tabular}{ll}
\hline
\textbf{Training Setting} & \textbf{Configuration} \\
\hline
Epochs & 3 \\
Batch size & 4 \\
Learning rate & 1e-4 \\
Optimizer & paged\_adamw\_32bit \\
Gradient accumulation steps & 2 \\
Warmup steps & 10 \\
Max sequence length & 512 tokens (8,192 tokens for OpenCoder-LLM) \\
Hardware & NVIDIA H100 (80GB) \\
\hline
\end{tabular}
\vspace{.7em}
\caption{\textbf{Training hyperparameters used across all experiments.}}
\label{tab:hyperparams}
\end{table}

Data was collated using the {\tt DataCollatorForCompletionOnlyLM} from the TRL library, computing loss only on the response portion. We deploy all models to the Hugging Face Hub with standardized nomenclature indicating the base model, dataset, training approach, and key hyperparameters.

\subsection{Few-shot Task Embedding Configuration}
\label{appendix:fewshot}
For the few-shot task embedding experiments described in Section \ref{sec:task-embed}, we used a configuration tailored for limited data scenarios. We randomly sampled 20 examples from each domain's held-out data that was not used in any of our main experiment splits.

\begin{table}[h]
\small
\centering
\begin{tabular}{ll}
\hline
\textbf{Parameter} & \textbf{Value} \\
\hline
Examples per domain & 20 (randomly sampled) \\
Learning rate & 3.3e-3 \\
Batch size & 1 \\
Epochs & 5 \\
\hline
\end{tabular}
\vspace{.7em}
\caption{Hyperparameters for few-shot task embedding experiments.}
\label{tab:fewshot_hyperparams}
\end{table}

We maintained the same LoRA configuration as in our main experiments. The higher learning rate and increased number of epochs compensate for the limited training data while the smaller batch size allows for more frequent parameter updates. After fine-tuning, we computed Delta Activations using identical probe datasets as in our main experiments to ensure direct comparability between few-shot task embeddings and full model embeddings in the embedding space.

\subsection{Delta Meaning implementation}

\label{appendix:delta-meaning}

 Here we provide implementation details for \textit{Delta Meaning}, our adaptation of Meaning Representations~\citep{liu2024meaning} to the Delta framework. This extension enables model embeddings across heterogeneous backbones, where direct activation comparisons are infeasible.

\paragraph{Meaning representations.}
Given a probe prompt $x$, we first sample $n$ continuations $\{s_1, \dots, s_n\}$ from the base model (temperature = 1.0). For any finetuned model $f$, we then score each continuation $s_i$ by computing its inverse perplexity under $f$:
\[
m_f(x)_i = \exp\!\left(- \frac{1}{|s_i|} \sum_{t=1}^{|s_i|} \log p_f(s_{i,t} \mid s_{i,<t}, x) \right).
\]
This produces an $n$-dimensional ``meaning vector'' $m_f(x) \in \mathbb{R}^n$ for each prompt $x$.

\paragraph{Delta aggregation.}
For a finetuned model $f$ and its base $f_{\text{base}}$, we define the \textit{Delta Meaning} on prompt $x$ as the difference between their meaning vectors:
\[
\Delta_f(x) = m_f(x) - m_{f_{\text{base}}}(x).
\]
Aggregating across all prompts $x \in D_{\text{probe}}$ yields the final model embedding:
\[
v_f = \frac{1}{|D_{\text{probe}}|} \sum_{x \in D_{\text{probe}}} \Delta_f(x).
\]

\textbf{Hyperparameters.}
In our experiments, we set $n$ to be 20. Larger $n$ provides more informative embeddings but requires proportionally more forward passes, as each continuation must be scored by both the base and finetuned models. Despite this, the dimensionality of Delta Meaning remains extremely compact compared to weight- or logit-based alternatives. Importantly, because any model can evaluate the probability of a given text sequence, Delta Meaning embeddings are naturally architecture-agnostic, allowing us to cluster finetuned models drawn from multiple backbones.

\newpage

\newpage
\section{Additional Analysis}

\subsection{t-SNE Visualization on more backbones}

\begin{figure}[h]
    \centering
    \includegraphics[width=\textwidth]{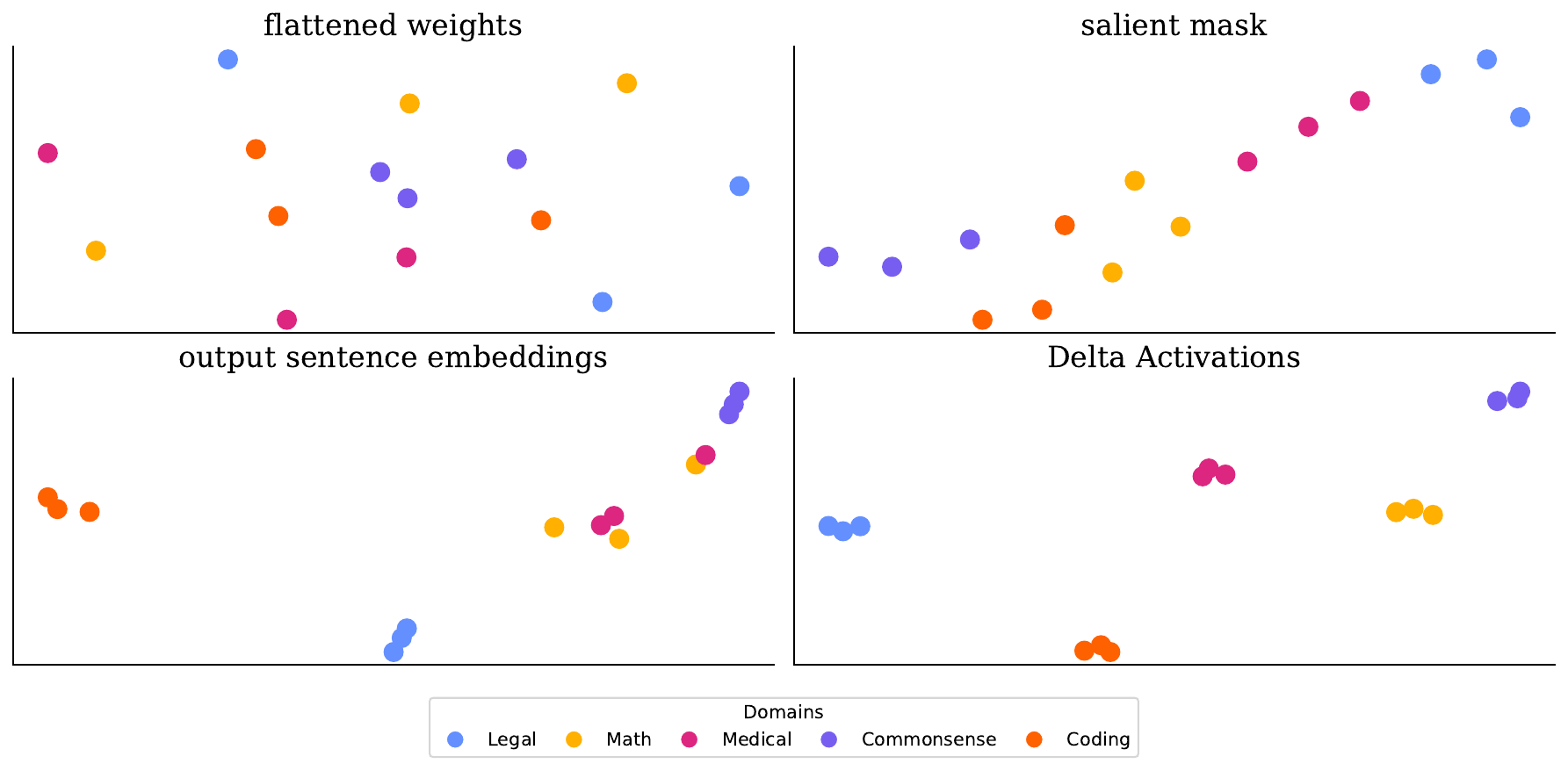}
    \caption{\textbf{t-SNE visualization of different embedding spaces (LLaMA).}}
    \label{fig:tsne-comparison-llama}
\end{figure}

\begin{figure}[h]
    \centering
    \includegraphics[width=\textwidth]{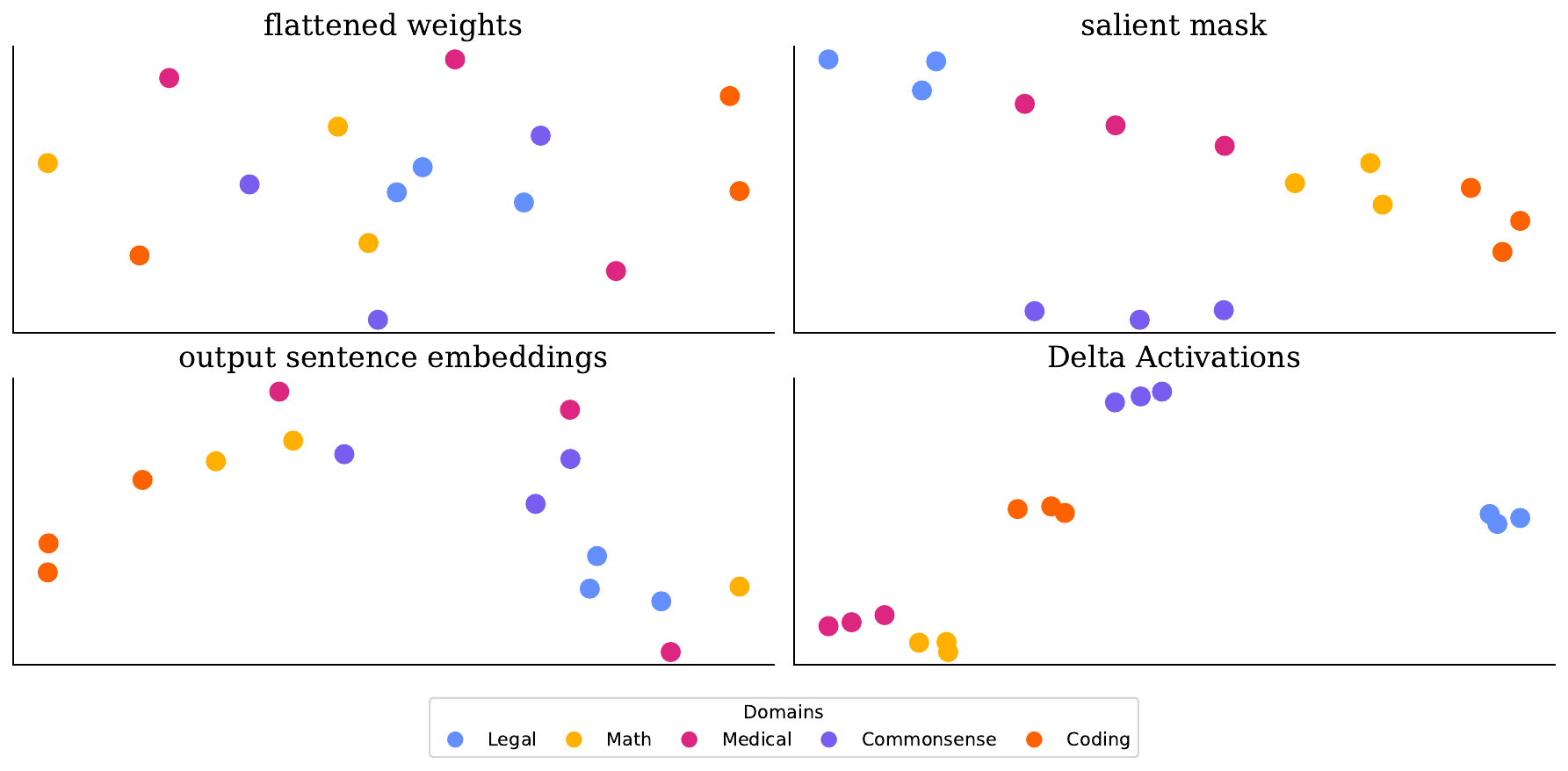}
    \caption{\textbf{t-SNE visualization of different embedding spaces (Qwen).}}
    \label{fig:tsne-comparison-qwen}
\end{figure}

We show visualization of the experiments conducted in Section \ref{sec:model_embed} on LLaMA and Qwen in Figure~\ref{fig:tsne-comparison-llama} and Figure~\ref{fig:tsne-comparison-qwen} respectively. Over different backbones, the visualization consistently shows the superiority of Delta Activations in forming cleanly-separated clusters.

\newpage

\subsection{Full Results for Additive Properties experiment}
To better understand the additive nature of Delta Activations, we extend the experiment from Section~\ref{sec:model_embed} to cover all ten domain pairs. In each case, we compare the Delta Activation vector from a model trained on the mixed dataset against those from models trained individually on each domain, as well as their vector sum. This comprehensive table demonstrates that the summed Delta Activations consistently better approximate the mixed-model embedding, reinforcing the additive property of Delta Activations.
\begin{table}[ht]
    \centering
    \small
    \begin{tabular}{ccccc|ccc}
    \toprule
    \textbf{Math} & \textbf{Commonsense} & \textbf{Code} & \textbf{Medical} & \textbf{Legal} & \textbf{Mixed v. D1} & \textbf{Mixed v. D2} & \textbf{Mixed v. Sum} \\
    \midrule
    \checkmark & & \checkmark & & & .703 & .270 & \textbf{.726} \\
    \checkmark & \checkmark & & & & .577 & .484 & \textbf{.649} \\
    & \checkmark & \checkmark & & & .631 & .283 & \textbf{.653} \\
    & & & \checkmark & \checkmark & .407 & .675 & \textbf{.695} \\
    & \checkmark & & \checkmark & & .359 & .662 & \textbf{.677} \\
    \checkmark & & & \checkmark & & .760 & .697 & \textbf{.811} \\
    & & \checkmark & \checkmark & & .462 & .581 & \textbf{.693} \\
    & \checkmark & & & \checkmark & .649 & .659 & \textbf{.763} \\
    \checkmark & & & & \checkmark & .522 & .610 & \textbf{.669} \\
    & & \checkmark & & \checkmark & .445 & .507 & \textbf{.587} \\
    \end{tabular}
    \vspace{.7em}
    \caption{\textbf{Additive property (full).} The sum of Delta Activations on models finetuned separately on two datasets aligns well with Delta Activations on the model finetuned on two datasets mixed together.}
    \label{tab:additive_full}
\end{table}

\subsection{Full finetuning}

\begin{figure}[h]
    \centering
    \includegraphics[width=.7\textwidth]{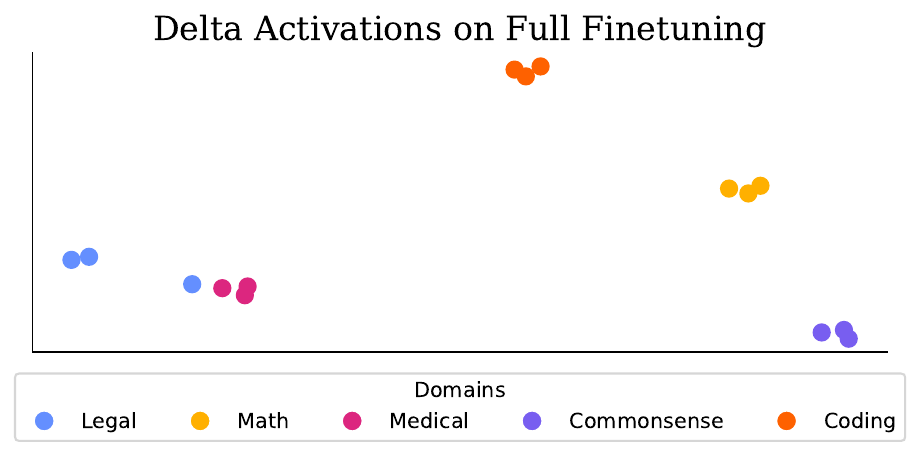}
    \caption{\textbf{t-SNE visualization of full finetuning.}}
    \label{fig:tsne-comparison-fft}
\end{figure}

We conduct experiments in Section~\ref{sec:model_embed} on \textsc{LLaMA-3.1-8B} with the model pool trained using full finetuning instead of LoRA. This experiment yields a silhouette score of \textbf{0.63}, which confirms that Delta Activations provide clear clustering regardless of finetuning methods. Visualization is shown in Figure~\ref{fig:tsne-comparison-fft}.

\newpage

\textbf{Intra-domain clustering.} We conduct experiments to test whether Delta Activations provide sufficient signal to differentiate models trained on different sub-expertises within medical and coding domains. 

We partition each domain into multiple sub-expertise splits. For medical, we create finetuned 8 models: disease databases \citep{freedomintelligence2024disease} (2 splits), disease-symptoms \citep{mohamedahmed2024diseasesymptoms} (2 splits), MedQA-USMLE \citep{jin2020disease} (2 splits), and MedMCQA \citep{pmlr-v174-pal22a} (2 splits). For coding, we create 6 models covering C++ (2 splits), Python (2 splits), and Java (2 splits) from CodeChef \citep{codechef}, Codeforces \citep{quan2025codeelobenchmarkingcompetitionlevelcode}, and Evol-Instruct \citep{luo2023wizardcoder} datasets. We compute Delta Activations using both generic instruction prompts and domain-specialized prompts.

Table~\ref{tab:intra-domain} shows that generic prompts achieve a silhouette score of 0.657 for medical sub-expertise, outperforming specialized medical prompts (0.533). For coding, both prompts perform comparably (0.668 vs 0.674). The generic probe maintains sufficient resolution to form correct clusters of sub-expertises within domains.

\begin{table}[ht]
    \centering
    \small
    \begin{tabular}{crr}
    \toprule
    \textbf{Domain} & \textbf{Generic prompt} & \textbf{Specialized prompt} \\
    \midrule
    Medical & \textbf{.657} & .533 \\
    Coding & .668 & \textbf{.674} \\
    \end{tabular}
    \vspace{.7em}
    \caption{\textbf{Intra-domain clustering.} Silhouette scores for sub-expertise clustering within domains.}
    \label{tab:intra-domain}
\end{table}


\newpage

\section{Prompt Templates}
\label{appendix:prompts}
Table~\ref{tab:prompt_templates}, Table~\ref{tab:one_sentence_templates}, and Table~\ref{tab:one_word_templates} list all prompt templates used in probe dataset experiments in Section~\ref{sec:understanding}. For other experiments, we use the first five prompts in Table \ref{tab:prompt_templates}.

\begin{table}[ht]
    \centering
    \small
    \begin{tabular}{|c|p{12cm}|}
        \hline
        \textbf{ID} & \textbf{Prompt Template} \\
        \hline
        1 & Below is an instruction that describes a task. Write a response that appropriately completes the request. \\
        \hline
        2 & The task described below requires a response that completes the request accurately. \\
        \hline
        3 & Below is a description of a task. Provide a response that aligns with the requirements. \\
        \hline
        4 & The following instruction outlines a task. Generate a response that meets the specified request. \\
        \hline
        5 & You are given an instruction and input. Write a response that completes the task as requested. \\
        \hline
        6 & You are provided with a task instruction and input. Write a response that fulfills the described requirements. \\
        \hline
        7 & Here is an instruction and its associated input. Complete the task with an appropriate response. \\
        \hline
        8 & Below is a task along with its context. Write a response that matches the requirements. \\
        \hline
        9 & The following is a description of a task and its input. Generate a response that fulfills the request. \\
        \hline
        10 & An outlined task is provided along with its input. Write a response that satisfies the given instruction. \\
        \hline
        11 & Given the following instruction, generate a suitable response that fulfills the request. \\
        \hline
        12 & The task described below requires a response that completes the request accurately. \\
        \hline
        13 & Below is a description of a task. Provide a response that aligns with the requirements. \\
        \hline
        14 & The following instruction outlines a task. Generate a response that meets the specified request. \\
        \hline
        15 & You are given an instruction and input. Write a response that completes the task as requested. \\
        \hline
        16 & Here is an instruction and its associated input. Create a response that properly addresses the request. \\
        \hline
        17 & Below is a task description. Provide an appropriate response that matches the input. \\
        \hline
        18 & An instruction and input are provided. Write a response that accurately completes the task. \\
        \hline
        19 & The following is an instruction that describes a task. Write a response that correctly satisfies the request. \\
        \hline
        20 & Below is an outlined task. Respond with a completion that fits the instruction and input given. \\
        \hline
    \end{tabular}
    \vspace{.7em}
    \caption{List of paraphrased prompt templates used in our experiments.}
    \label{tab:prompt_templates}
\end{table}

\begin{table}[ht]
    \centering
    \begin{minipage}{0.6\textwidth}
        \centering
        \small
        \begin{tabular}{|c|p{6cm}|}
            \hline
            \textbf{ID} & \textbf{Prompt Template} \\
            \hline
            1 & Instruction: Please provide a response. Input: Input. \\
            \hline
            2 & Please perform the following task.  \\
            \hline
            3 & Complete the instruction.  \\
            \hline
            4 & Provide the appropriate response.  \\
            \hline
            5 & Here is the text. Response:  \\
            \hline
        \end{tabular}
        \vspace{.7em}
        \caption{One-sentence paraphrased prompt templates.}
        \label{tab:one_sentence_templates}
    \end{minipage}%
    \hfill
    \begin{minipage}{0.35\textwidth}
        \centering
        \small
        \begin{tabular}{|c|p{3cm}|}
            \hline
            \textbf{ID} & \textbf{Prompt Template} \\
            \hline
            1 & Response: \\
            \hline
            2 & Answer: \\
            \hline
            3 & Explanation: \\
            \hline
            4 & Solution: \\
            \hline
            5 & Discussion: \\
            \hline
        \end{tabular}
        \vspace{.7em}
        \caption{One-word paraphrased prompt templates.}
        \label{tab:one_word_templates}
    \end{minipage}
\end{table}

\section{Broader Impact}
\label{appendix:impact}
The proposed Delta Activations method facilitate efficient reuse of fine-tuned models by providing an embedding to encode the finetuned model's behaviors and capability. This reduces redundant training, cutting energy costs and promoting sustainable AI practices. Furthermore, it encourages broader public sharing of fine-tuned models by offering clear documentation of their capabilities, accelerating research and collaboration.

However, expanding public model hubs also introduces risks, as low-quality or adversarial models could contaminate the pool. This highlights the need for careful curation to maintain reliability and safety in open model ecosystems.

\newpage



\end{document}